\documentclass[iicol,sn-mathphys,Numbered]{sn-jnl}

\usepackage{graphicx}%
\usepackage{multirow}%
\usepackage{amsmath,amssymb,amsfonts}%
\usepackage{amsthm}%
\usepackage{mathrsfs}%
\usepackage[title]{appendix}%
\usepackage{xcolor}%
\usepackage{textcomp}%
\usepackage{manyfoot}%
\usepackage{booktabs}%
\usepackage{algorithm}%
\usepackage{algorithmicx}%
\usepackage{algpseudocode}%
\usepackage{listings}%
\usepackage[numbers]{natbib} 
\usepackage{array}
\usepackage[caption=false,font=normalsize,labelfont=sf,textfont=sf]{subfig}
\usepackage{stfloats}
\usepackage{url}
\usepackage{verbatim}
\newcommand{\cmark}{\color{black}{\ding{51}}}
\newcommand{\xmark}{\color{black}{\ding{55}}}
\usepackage{pifont}
\usepackage{hyperref}
\usepackage{adjustbox}

\begin{document}

\title[Article Title]{DAD++: Improved 
Data-free Test Time Adversarial Defense}


\author[1,2]{\fnm{Gaurav Kumar} \sur{Nayak}}\email{gauravkumar.nayak@ucf.edu}\equalcont{Work done by authors at Indian Institute of Science, Bangalore.}

\author[2]{\fnm{Inder} \sur{Khatri}}\email{inderkhatri999@gmail.com}

\author[2]{\fnm{Shubham} \sur{Randive}}\email{randiveshubham3@gmail.com}
\equalcont{Work done by authors at Indian Institute of Science, Bangalore.}

\author[2,3]{\fnm{Ruchit} \sur{Rawal}}\email{rrawal@mpi-sws.org}
\equalcont{Work done by authors at Indian Institute of Science, Bangalore.}

\author*[2]{\fnm{Anirban} \sur{Chakraborty}}\email{anirban@iisc.ac.in}

\affil[1]{\orgdiv{Centre for Research in Computer Vision}, \orgname{University of Central Florida}, \orgaddress{\country{USA}}}

\affil[2]{\orgdiv{Department of Computational and Data Sciences}, \orgname{Indian Institute of Science}, \orgaddress{\city{Bangalore},\postcode{560012},\country{India}}}

\affil[3]{\orgname{Max Planck Institute for Software Systems}, \orgaddress{\city{Saarbrücken}, \country{Germany}}}


\abstract{With the increasing deployment of deep neural networks in safety-critical applications such as self-driving cars, medical imaging, anomaly detection, etc., adversarial robustness has become a crucial concern in the reliability of these networks in real-world scenarios. A plethora of works based on adversarial training and regularization-based techniques have been proposed to make these deep networks robust against adversarial attacks. However, these methods require either retraining models or training them from scratch, making them infeasible to defend pre-trained models when access to training data is restricted. To address this problem, we propose a test time Data-free Adversarial Defense (DAD) containing detection and correction frameworks. Moreover, to further improve the efficacy of the correction framework in cases when the detector is under-confident, we propose a soft-detection scheme 
(dubbed as ``DAD++"). We conduct a wide range of experiments and ablations on several datasets and network architectures to show the efficacy of our proposed approach. Furthermore, we demonstrate the applicability of our approach in imparting adversarial defense at test time under data-free (or data-efficient) applications/setups, such as Data-free Knowledge Distillation and Source-free Unsupervised Domain Adaptation, as well as Semi-supervised classification frameworks. We observe that in all the experiments and applications, our DAD++ gives an impressive performance against various adversarial attacks with a minimal drop in clean accuracy. The source code is availabe \href{https://github.com/vcl-iisc/Data-free-Defense-at-test-time}{here}.}

\keywords{Adversarial Defense, Absence of Training Data, Test time defense, Fourier Transformation, Semi-Supervised Classification, Data-free Knowledge Distillation, Source-free Domain Adaptation}



\maketitle

\section{Introduction}
\label{sec:intro}
Deep learning, a subset of machine learning that involves the use of deep neural networks, has achieved impressive results in a wide range of fields, including computer vision~\cite{Voulodimos2018DeepLF,Rawat2017DeepCN}, natural language processing~\cite{Manning1999FoundationsOS,Jurafsky2000SpeechAL}, and speech recognition~\cite{nassif2019speech,Hannun2014DeepSS}. Within the field of computer vision, deep learning models have achieved state-of-the-art performance on tasks such as image classification~\cite{He2015DeepRL,Rawat2017DeepCN}, object detection~\cite{ren2015faster,faster-nips-2015}, face recognition~\cite{yang2021attacks,vakhshiteh2021adversarial}, and many more~\cite{LeCun2015DeepL}. These advances have resulted in various practical applications, including self-driving cars~\cite{Badue2021SelfDrivingCA}, security systems~\cite{faceid}, anomaly detectors~\cite{Chandola2009AnomalyDA}, etc. Despite their impressive performance, deep learning models are often vulnerable to adversarial attacks, which are human imperceptible perturbations to input data crafted with the intention of deceiving deep learning models. If misused, adversarial attacks can lead to the manipulation of deep networks and pose a threat to their real-world deployment~\cite{kurakin2018adversarial}. As an example, adversarial vulnerabilities in self-driving cars can manipulate the vehicle's decision-making processes and pose a severe risk to the safety of passengers and others on the road.

Over the past several years, researchers have made substantial efforts to improve the robustness of deep neural networks against adversarial attacks~\cite{akhtar2018threat,moosavi2016deepfool}. These efforts can be divided into two main categories: Adversarial training~\cite{liu2021longterm,pmlr21xu} and Non-adversarial training~\cite{pang2017towards,higashi2020detection} approach. Adversarial training involves training deep networks on adversarial examples to improve their ability to yield correct predictions against adversarial samples. However, this can be time-consuming due to adversarial examples generation in each batch, and further may require additional computational resources and memory. Additionally, a model trained on one attack using adversarial training may not effectively defend against other adversarial attacks. Thus, it may require retraining to adapt to new attacks. Non-adversarial training approaches, on the other hand, aim to enhance the robustness of deep neural networks to adversarial attacks without training the target model on adversarial examples, using techniques such as data augmentation~\cite{eghbal2020data}, regularization~\cite{jakubovitz2018improving}, model ensembling~\cite{croce2020reliable}, and denoising network methods ~\cite{Mustafa2019ImageSA,Ding2021DelvingID}. These approaches are often less compute and time-intensive.

The most crucial disadvantage of these adversarial defense frameworks is their heavy reliance on a large and diverse dataset for training. It has been observed that achieving a decent adversarial accuracy for a deep neural network demands substantially more training data when compared to the amount of data required for obtaining high clean accuracy for the same model~\cite{schmidt2018adversarially}. This can be a significant challenge in real word scenarios, where access to training data may be limited or restricted for various reasons, including concerns about data privacy and proprietary rights. Companies may also be hesitant to share their data due to competitive concerns and the expense of preparing it for use in machine learning models. This limitation can be especially problematic when no training data is provided, and only a trained model is available. Moreover, existing adversarial defense approaches often require retraining by assuming access to training data, making them infeasible to provide robustness to any off-the-shelf pre-trained classifiers in the absence of training data. 

To address the limitation of current approaches, we propose an adversarial defense framework named ‘
DAD++’. We present a strategy that involves detecting and correcting adversarial input samples at test time. we train a classifier on arbitrary data to detect adversarial samples in the adversarial test set. To account for the domain shift between the arbitrary data and target test samples, we frame the detection problem as an unsupervised domain adaptation. We further utilize a source-free unsupervised domain adaptation technique to reduce dependency on arbitrary data to use any pre-trained detector. Our correction method is inspired by how humans ignore inputs outside a certain frequency range ~\cite{Wang2020TowardsFE}. We use this insight to mitigate the effects of adversarial attacks, which often disrupt model predictions by manipulating high-frequency components of the input. Removing too many high-frequency components can lead to lower discriminability while leaving too many intact can allow adversarial attacks to persist. Thus, we propose a novel algorithm to find a balance between removing enough high-frequency components to prevent attacks while not removing so many as to reduce discriminability. Our framework does not require retraining of the target model, also it does not depend on target training dataset. It is also architecture agnostic, and we demonstrate its efficacy on various network architectures without modifying the trained model. Thus, it can be utilized for any pre-trained classifier in the absence of training data. 

We conduct multiple ablation studies on DAD++'s performance. Firstly, we evaluate the performance of DAD++ on a test set containing a mix of adversarial samples generated using different adversarial attacks. 
We also evaluate DAD++ on test sets with varying proportions of clean and adversarial samples. This allows us to analyze the efficacy of DAD++ when there is a relative imbalance in clean and adversarial images in the test set. Through these ablations, we observe that our detector is able to successfully detect adversarial images created using different adversarial attacks, which leads to an appreciable performance even against the different attacks. Also, we find that our detector achieves good accuracy when the test set contains an imbalanced proportion of clean and adversarial images.  


As our proposed approach is a test-time method that does not depend on the availability of training data labels, making it even suitable for providing robustness to label-efficient setups such as Semi-Supervised classification. While it is commonly recommended to train a model on a large labeled dataset to improve generalizability and reduce overfitting, this is generally infeasible due to the time-consuming and costly label annotation process. 
Therefore, semi-supervised~\cite{Sohn2020FixMatchSS, Zhang2021FlexMatchBS}  approaches are desirable, which can work with a small number of labeled samples by leveraging the abundantly available unlabeled samples. However, the robustness of these label-efficient approaches is a matter of concern. Recently, some works have attempted to use additional unlabeled data to improve the adversarial robustness through semi-supervised training~\cite{carmon2019unlabeled,alayrac2019labels,zhai2019adversarially}. Despite their capability to improve the adversarial robustness of semi-supervised target frameworks, they still require retraining of the target model with the help of a reasonable amount of labeled data. Our proposed approach can overcome these limitations and is effective in defending the pre-trained semi-supervised frameworks without any retraining, irrespective of the training strategy.




One of the limiting properties of neural networks is their requirement for deep architectures to achieve desirable performance, which makes their deployment for live applications challenging. Generally, deeper networks are not suitable for deploying on portable devices due to the hardware constraints, needing them to be lightweight. However, due to the less accuracy of lightweight models, their capability to solve real-world problems reduces. In such cases, Knowledge Distillation can aid in providing a good trade-off between performance and efficiency. Knowledge Distillation~\cite{chen2019data,Micaelli2019ZeroshotKT} is capable of transferring knowledge from larger, complex teacher models to smaller, more efficient student models for improving the student's performance with the help of training data. Recently, there have been developments in Zero-Shot Knowledge Distillation frameworks ~\cite{zskt-neurips-2019} 
that can distill knowledge from a teacher to a student model, even without training data. Although student models obtained using data-free methods are able to show impressive performance, incorporating adversarial robustness into the student model in these settings is challenging. Since our framework does not require training data, it can be easily used to defend the distilled student models obtained using any data-free distillation methods against various adversarial attacks.


Another important concern of the deep networks is their inability to sustain performance when used for the data samples from a domain (target) different from the one originally used for the training. Further, training a new model for the target domains requires labeled data, which can bring the heavy cost of annotating with it. To address such issues, Unsupervised Domain Adaptation~\cite{Liang2021DINEDA,Ahmed2021UnsupervisedMD,liang2020we} approaches are used to adapt the model using labeled data from the source domain and unlabeled data from the target domain. However, due to privacy and security concerns, source data is not always available, and only a pre-trained source model is available. In such cases, Source-free Domain Adaptation frameworks can be utilized~\cite{liang2020we}. However, obtaining an adversarial robust adapted model (i.e., target model robust against adversarial attacks) in these scenarios can be difficult, as no labels and even no unlabeled data are available from the source domain. Our proposed DAD++ framework is well-suited even for these situations, as it does not depend on any training data and can be effectively applied in these settings.


We summarize our contributions as follows:
\begin{itemize}
    \item We introduce a novel problem of Data-free Adversarial Defense (DAD++) at test time. Our proposed solution to this problem does not require any retraining, nor it requires any data to defend against adversarial attacks.
    \item Our approach consists of two steps: adversarial sample detection and correction. We present a new method of training the detector using source-free domain adaptation. Our correction module is inspired by human cognition ignoring high-frequency regions in the frequency domain. 
    \item We propose a novel ‘Soft Detection’ mechanism, which is able to deal with the uncertainty associated with the detector’s prediction. 
    \item We assess the effectivenss of DAD++ in various challenging scenarios, such as imbalanced data with varying proportions of clean and adversarial samples in the test set, a mix of samples created with different attacks in the adversarial set, and small test set sizes. 
    \item We show the efficacy of our approach (DAD++) for semi-supervised learning frameworks and also demonstrate the utility of DAD++ for two new applications (Zero-Shot Knowledge Distillation and Source-free Domain Adaptation).
\end{itemize}

Note that the framework for Data free defense at test time was originally introduced in our earlier conference paper~\cite{Nayak2022DADDA}.  All the other contributions are novel additions to this extended article. The framework that was initially introduced in the conference paper is referred as DAD, whereas the framework presented in this expanded article is referred as DAD++.

\section{Related Works}
\label{sec:related}
\textbf{Adversarial Detection and Robustness: } Adversarial detection methods aim to successfully detect adversarially perturbed images. These can be broadly achieved by using trainable-detector or statistical analysis based methods. The former usually involves training a detector network either directly on the clean and adversarial images in spatial~\cite{lee2018unified, ma2018lid, jan2017detect} / frequency domain~\cite{Harder2021SpectralDefenseDA} or on logits computed by a pre-trained classifier~\cite{jonathan2019introspect}. Statistical analysis based methods employ statistical tests like maximum mean discrepancy~\cite{kathrin2017detect} or propose measures~\cite{reuben2017detect, bin2021detect} to identify perturbed images. It is impractical to use the aforementioned techniques in our problem setup as they require training data to either train the detector (trainable detector based) or tune hyper-parameters (statistical analysis based). We tackle this by formulating the detection of adversarial samples as a source-free unsupervised domain adaptation setup wherein we can adapt a detector trained on arbitrary (source) data to unlabelled (target) data at test time. 

Adversarial training is one of the first methods proposed to make a model robust to adversarial perturbations. Szegedy~\cite{szegedy2013intriguing} first articulated adversarial training as augmenting the training data with adversaries to improve the robustness performance of a specific attack. Madry~\cite{madry2017towards} proposed the Projected Gradient Descent (PGD) attack, which when used in adversarial training provided robustness against a wide class of iterative and non-iterative attacks. 
Apart from adversarial training, other non-adversarial training based approaches primarily aim to regularize the network to reduce overfitting and achieve properties observed in adversarially trained models explicitly. Most notably, Jacobian Adversarially Regularized networks~\cite{chan2019jacobian} achieve adversarial robustness by optimizing the model’s jacobian to match natural training images.
In recent years, few works have proposed an approach to modify the input image to remove the adversarial perturbations instead of retraining the target model. One such work, ComDefend~\cite{Jia2018ComDefendAE} uses a compression-reconstruction deep neural network, which removes the adversarial perturbation by compressing the input image. They specifically compact the original high-bit input image to the low-bit representation to remove the redundant fine details and retain the dominant information. Further, this low-bit image is reconstructed using the reconstruction deep neural network by incorporating the fine details removed during the compression. Similarly, Mustafa et al~\cite{Mustafa2019ImageSA} use deep image super-resolution networks to remove the adversarial perturbations. They observed that super-resolution networks could bring off-the-manifold adversarial images to the manifold where clean images lie, and thus, super-resolution networks can aid in defending against adversarial attacks. Naseer et al.~\cite{Naseer2020ASA} proposed a generalizable Neural Representation purifier that purifies the adversarial images across different tasks. This purifier is trained using self-supervised perturbations, which makes it generalizable across the attacks. 

These methods can help in providing adversarial robustness to any off-the-shelf pre-trained models, as they do not involve retraining the target model. However, a significant drawback of these approaches is the heavy reliance on external priors in the form of large training data to train the purifier/denoiser networks. Further, their performance can drop catastrophically when utilized for the classifiers from the shifted domain in real-world setups. To overcome these limitations, Dai et al. proposed a Deep Image Prior~\cite{Ulyanov2017DeepIP} Driven Defense (DIPDefend)~\cite{Dai2020DIPDefendDI} to reconstruct the clean sample corresponding to the given adversarial sample. They use the PSNR value of the reconstructed image to decide the iteration at which reconstruction should stop. Once the rate of increase of PSNR value drops below a certain threshold the network stops refining the image. Further, Ding et al.~\cite{Ding2021DelvingID} improved upon DIPDefend by proposing a deep image prior based reconstruction network. The reconstruction network reliably detects on-boundary images in the decision space. Using on-boundary images, images lying in a clean manifold are constructed. The final image is constructed by stitching images that lie on a clean manifold. However, DIP-based approaches require hundreds of iterations to reconstruct an image, also these networks can process only one image at any given time. Because of these reasons, DIP based approaches are not viable for real-world applications. 

\textbf{Knowledge Distillation: }Knowledge distillation is a process by which the knowledge or expertise learned by a complex, high-performing machine learning model (called the teacher) is transferred to a simpler, lighter-weight model (called the student). Hinton et al.~\cite{hinton2015distilling} proposed one of the simplest methods for Knowledge Distillation by penalizing the student model to make its output similar to the soft output of the teacher model using KL divergence loss. Romero et al.~\cite{romero2014fitnets} extended this concept by attempting to match the intermediate features of the student model with those of the teacher model in order to transfer high-dimensional knowledge representations. Yim et al.~\cite{yim2017gift} 
proposed a framework for transferring relational knowledge from the teacher to the student, focusing on mimicking the relationships between the intermediate layers of the teacher model to improve the generalizability of the student. A major limitation of many knowledge distillation methods is their reliance on training data, which is not always available. To address this issue, there have been recent developments in data-free knowledge distillation techniques that do not require access to the original training data. One effective approach to data-free knowledge distillation involves using generative networks to learn proxy or synthetic data for transferring knowledge from one model to another. For example, Chen et al.~\cite{chen2019data} proposed a method called DAFL that treats the teacher network as a fixed discriminator and uses it to guide the generator network to produce training samples that it can predict with high confidence. Similarly, Micaelli et al.~\cite{Micaelli2019ZeroshotKT} used the generator in an adversarial manner to synthesize training samples on which the student network has not yet learned to make predictions that align with those of the teacher network.

Several research studies have attempted to incorporate adversarial robustness, or the ability to resist noise or malicious attacks, into student models through variations of knowledge distillation.  A common approach has been to combine knowledge distillation with adversarial training~\cite{goldblum2020adversarially,zhu2022reliable,zi2021revisiting}. 
Goldblum et al.~\cite{goldblum2020adversarially} introduced a method called ARD that forces the student model to mimic the teacher model’s predictions on clean samples while predicting noisy samples. Zhu et al.~\cite{zhu2022reliable} proposed a robust distillation framework called reliable IAD, which allows the student model to rely partially on the teacher model's predictions based on their accuracy. Zi et al.~\cite{zi2021revisiting} improved upon ARD by using the robust soft labels produced by a teacher model to learn from both clean and adversarial samples. . Other researchers have tried to transfer robustness from a pre-trained, robust teacher model to the student model without using adversarial samples. Chen et al.~\cite{chan2020thinks} proposed an approach called IGAM, where they made students mimic the input gradients of the adversarial robust teacher model by fooling the discriminator model. Similarly, Shao et al.~\cite{shao2021and} aimed to align the student model's gradient with that of the robust teacher model by adding a loss term that penalizes the student model based on the difference between the gradients. However, most of these methods for creating robust student networks require either adversarial data or robust teacher networks. In contrast, our proposed Data-free Adversarial Defense (DAD++) method provides adversarial robustness to distilled student models without access to training data, regardless of the distillation method used.\\

\textbf{Unsupervised Domain Adaptation:} Unsupervised Domain Adaptation (UDA) aims to fine-tune models trained on source data to accurately predict labels for target data without labels. Traditionally, source and target data were required to be available for the adaptation process. However, some recent research works have focused on eliminating the need for source training data, allowing a pre-trained model from the source domain to be transferred to a target domain using the unlabeled target data\cite{liang2020we,Ahmed2021UnsupervisedMD,Liang2021DINEDA}. Liang et al.~\cite{liang2020we} introduced one of the first approaches in this direction called SHOT, which uses the source model’s classifier module (hypothesis) for the target domain as it is and adapts the feature extractor of the source model using only unlabeled target data. The feature extractors are fine-tuned using the combination of information maximization loss and self-supervised pseudo-labeling to align the representations of the target domain with the source hypothesis without the need for explicit alignment. However, SHOT is limited because it is designed to use a single source model for adaptation and cannot utilize knowledge from more than one model when multiple source models trained on different domains are available. To address this, Ahmed et al.~\cite{Ahmed2021UnsupervisedMD} proposed an algorithm called DECISION, which automatically combines multiple source models with appropriate weights to ensure that the resulting model performs at least as well as the best source model. Additionally, Liang et al.~\cite{Liang2021DINEDA} recognized the challenge of adapting black box models without access to source data and proposed a two-step approach called DINE. DINE first distills knowledge from the source predictor to a customized target model and then fine-tunes the distilled model on unlabeled target data to fit the target domain better, allowing it to adapt to the target data effectively.

Transferring the adversarial robustness of a source model to a target domain during model adaptation is challenging. One way to do this is through adversarial training on labeled source data during domain adaptation. However, this may not be effective as the boundaries of adversarial data may not align across the two domains. Awais et al.~\cite{awais2021adversarial} recently attempted to impart robustness to models during UDA without requiring adversarial training. They proposed a method that utilizes the intermediate representations learned by robust ImageNet models to improve the robustness of UDA models by aligning the features of the UDA model with the robust features learned by pre-trained ImageNet models during domain adaptation training. While this approach improves the robustness of UDA models, it still requires access to source data, making it unsuitable for source-free domain adaptation methods. In contrast, the proposed DAD++ framework can provide adversarial robustness to networks trained using source-free domain adaptation in a test-time defense manner, making it more applicable to situations where source data is unavailable. 

\textbf{Semi-Supervised Learning:}
Semi-supervised learning (SSL) is a machine learning paradigm that utilizes both labeled and unlabeled data for training, where the goal is to improve model accuracy compared to using only labeled data. SSL is especially useful when it is costly or time-consuming to obtain labeled data, but many unlabeled data samples are available. SSL methods can be grouped into several categories: generative methods, consistency regularization methods, graph-based methods, pseudo-labeling methods, and hybrid methods~\cite{yang2022survey}. Hybrid methods combine ideas from multiple approaches and are considered the most advanced and have demonstrated strong performance. Fixmatch~\cite{Sohn2020FixMatchSS} is one of the simplest Hybrid SSL methods that involves generating pseudo-labels for unlabeled images using the model's predictions. The method first applies weak augmentation to the unlabeled images and uses the model's predictions to generate pseudo-labels for these images. A pseudo-label is only retained for a given image if the model produces a high-confidence prediction. The model is then trained to predict the pseudo-label when given a strongly-augmented version of the same image. Another work, Flexmatch~\cite{Zhang2021FlexMatchBS}, which is an improvement on Fixmatch, introduces a technique called Curriculum Pseudo Labeling (CPL). CPL leverages unlabeled data according to the model's learning status and has been shown to significantly improve the performance and convergence speed of SSL algorithms that use thresholds. CPL is simple to implement and almost cost-free, making it a useful addition to SSL approaches like Fixmatch.

Motivated by SSL approaches, some works have attempted to utilize unlabeled data to improve the robustness of deep networks. Carmon et al. proposed the Robust Self Training (RST) mechanism, which, just like the standard SSL methods, utilized the pseudo-labels from the model trained in a supervised manner for training on unlabeled data. They observed that the adversarial performance on the standard supervised learning task could be improved by using the abundantly available unlabeled data.   

\textbf{Frequency Domain: }
Wang~\cite{wang2020high} in their work demonstrated that, unlike humans, CNN relies heavily on high-frequency components (HFC) of an image. Consequently, perturbations in HFC cause changes in model prediction but are imperceptible to humans. More recently, Wang~\cite{wang2020towards} showed that many of the existing adversarial attacks usually perturb the high-frequency regions and proposed a method to measure the contribution of each frequency component towards model prediction. Taking inspiration from these recent observations we propose a novel detection and correction module that leverages frequency domain representations to improve adversarial accuracy (without plunging clean accuracy) at test-time. The next section explains important preliminaries followed by the proposed approach in detail.

\section{Preliminaries}
\label{sec:prelims}
\textbf{Notations}: The target model, denoted by $T_{m}$, is pre-trained on a training dataset referred to as $D_{train}$. The complete target dataset is composed of both the training dataset and a test dataset, which is referred to as $D_{target} = {D_{train}, D_{test}}$. However, it is important to note that we do not have access to the training dataset $D_{train}$, but we do have access to the trained model $T_{m}$. The test dataset, $D_{test}={x_{i}}{i=1}^{N}$, contains N unlabelled test samples. Additionally, we consider a set of adversarial attacks represented by $A{attack}= {A_{j}}{j=1}^{K}$ where K is the total number of different attacks. The $i^{th}$ test sample, $x_{i}$, can be perturbed by any attack in the set of adversarial attacks $A_{attack}$, resulting in an adversarial sample denoted as $x_{i}^{'}$.

We refer to any dataset that is distinct from the target dataset $D_{target}$ as an arbitrary dataset, denoted by $D_{arbitrary}={(x_{iA}, y_{iA})}{i=1}^{M}$ that comprises of $M$ labeled samples. The model $S{m}$ is trained on the arbitrary dataset $D_{arbitrary}$. The adversarial sample that corresponds to $x_{iA}$ is denoted as $x_{iA}^{'}$. It is obtained when the trained model $S_{m}$ is attacked by any attack $A_{j} \in A_{attack}$. 

The logits predicted for the $i^{th}$ sample from the test dataset $D_{test}$ and the arbitrary dataset $D_{arbitrary}$ by the target model $T_{m}$ and the model $S_{m}$ respectively are denoted as $T_{m}(x_{i})$ and $S_m(x_{iA})$. The softmax function is represented by $soft()$. The label predicted by the target model $T_{m}$ and the model $S_{m}$ for the $i^{th}$ sample are denoted by $label(T_{m}(x_{i}))=\mathrm{argmax}(soft(T_{m}(x_{i})))$ and $label(S_{m}(x_{iA}))=\mathrm{argmax}(soft(S_{m}(x_{iA})))$ respectively. The complete set of adversarial samples that fool the target model $T_{m}$ and the model $S_{m}$ are denoted by $A_{test}$ and $A_{arbitrary}$ respectively, such that $x_{i}^{'} \in A_{test}$ and $x_{iA}^{'} \in A_{arbitrary}$. The set of layers used for adversarial detection are represented by $L_{advdet}$.

We denote the combination of the dataset $D_{test}$ and $A_{test}$ as $D_t$, and the combination of the dataset $D_{arbitrary}$ and $A_{arbitrary}$ as $D_s$. The $i^{th}$ sample in $D_{t}$ is denoted as $d_{i}$. Finally, the width and height of the image are denoted by $w$. For all the experiments width and height of the images are the same.

The Fourier transform and inverse Fourier transform are represented by $F(.)$ and $F^{-1}(.)$, respectively. The frequency component of a sample is designated as $f_{i}$. The low-frequency component (LFC) and high-frequency component (HFC) of a sample $f_{i}$ which are separated by a radius ($r$) are represented by $fl_{ir}$ and $fh_{ir}$, respectively.

\textbf{Naming Conventions}:  An incorrect detection of a clean image as adversarial is referred to as a false positive, and an incorrect detection of an adversarial image as clean is referred to as a false negative. The baseline accuracy reported in the tables is the accuracy of the target model without adversarial defense. The accuracy of the target model is reported as Clean accuracy/Adversarial accuracy, where clean accuracy is the accuracy of the target model on clean samples and adversarial accuracy is the accuracy of the target model on adversarial samples. Correction accuracy $\textbf{(Co.A)}$ is the accuracy of the target model when used in combination with a only correction module. In this case, all adversarial images are passed to the correction module, while clean images are directly fed to the target model. Therefore, the clean accuracy with only the correction module is the same as the baseline clean accuracy. Combined accuracy $\textbf{(Cb.A)}$ is the accuracy of the target model with the complete DAD++ defense. The combined clean accuracy is the clean accuracy of the target model with the DAD++ defense, and the combined adversarial accuracy is the adversarial accuracy with the DAD++ defense. Since the detector will have some false positives, the combined clean accuracy will always be less than the baseline clean accuracy. Similarly, due to false negatives, the combined adversarial accuracy will always be less than the correction accuracy. The accuracy of a target detector $\textbf{(TD.A)}$ is reported in terms of clean accuracy and adversarial accuracy. Clean accuracy denotes the percentage of clean samples that are correctly classified as clean, while adversarial accuracy denotes the percentage of adversarial samples that are correctly identified as adversarial. \textbf{Note}: For brevity, we use the terms 'model', and 'target model' interchangeably to refer to the target model. Similarly, we refer to the target dataset simply as 'dataset' and the target detector as 'detector'.

\begin{figure*}[htp]
\centering
\centerline{\includegraphics[width=\textwidth]{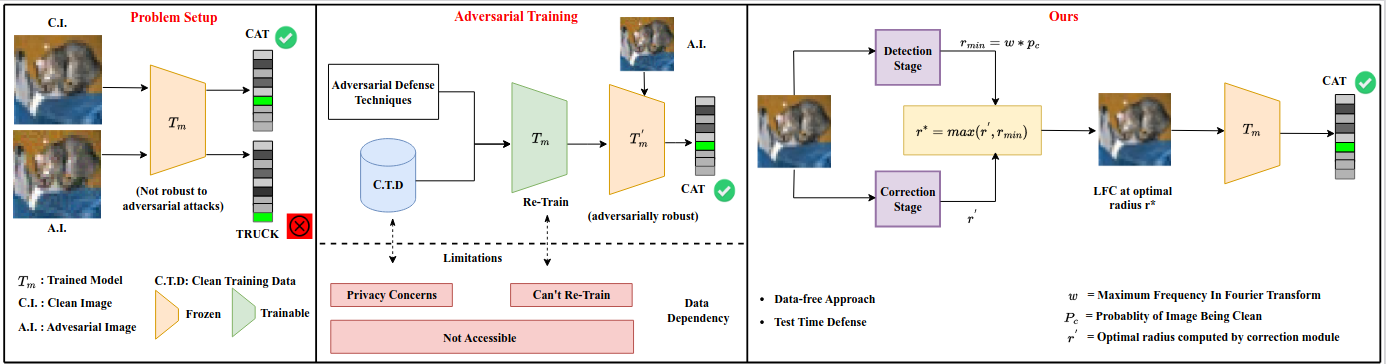}}
\caption{ \footnotesize Comparison of existing approaches and ours. Traditional approaches fail to handle data privacy scenarios, Our method provides data-free robustness against adversarial attacks at test time without even retraining the pre-trained non-robust model $T_{m}$. Our correction module utilizes a soft detection score to denoise the clean and adversarial images.}
\label{fig:overview}
\end{figure*}
\textbf{Adversarial noise}: Any adversarial attack $A_{j}$ that is included in the set of attacks $A_{attack}$ can fool a pre-trained network $T_{m}$ by altering the label prediction. An attack $A_{j}$ on the $i^{th}$ image $x_{i}$ generates an adversarial image $x_{i}^{'}$ such that $label(T_{m}(x_{i})) \neq label(T_{m}(x_{i}^{'}))$. To create $x_{i}^{'}$, the $i^{th}$ image $x_{i}$ is perturbed by an imperceptible adversarial noise $\delta$ such that $\left\lVert \delta \right\rVert$ is within a limit of $\epsilon$. We limit the perturbations within the $l_{\infty}$ ball of radius $\epsilon$.

\textbf{Unsupervised Domain Adaptation (UDA)}: A classifier $F_{s}$ is trained on a labeled dataset $D_{s}$ which comes from distribution $S$. The dataset $D_{t}$ which is unlabeled, belongs to a different distribution $T$. UDA methods aim to reduce the disparity between $S$ and $T$ with the goal of obtaining an adapted classifier $F_{t}$ using $F_{s}$ that can predict labels on the unlabeled samples from $D_{t}$. If we assume that $D_{s}$ is not available for adaptation, this problem is referred to as source-free UDA~\cite{liang2020we, li2020model, kurmi2021domain}.

\textbf{Fourier Transform (FT)}: The Fourier Transform (FT) is a method used in image processing to change the input image from the spatial domain to the frequency domain~\cite{bracewell1986fourier}. For any $i^{th}$ image $x_{i}$, its FT is defined as $F(x_{i}) = f_{i}$. At a specific radius $r$,
\begin{equation}
\begin{aligned}
fl_{ir} = LFC(f_{i}, r)\
fh_{ir} = HFC(f_{i}, r)
\end{aligned}
\end{equation}
The inverse Fourier Transform (IFT) helps to convert the image back to the spatial domain from the frequency domain. Thus, we have:
\begin{equation}
\begin{aligned}
xl_{ir} = F^{-1}(fl_{ir}) \
xh_{ir} = F^{-1}(fh_{ir})
\end{aligned}
\end{equation}
where $xl_{ir}$ and $xh_{ir}$ are the LFC and HFC of the $i^{th}$ image in the spatial domain.

\vspace{-0.1in}
\section{Proposed Approach}
\label{sec:proposed}
\textbf{Test-time Adversarial Defense Set up}: Our goal is to make a pre-trained target model $T_{m}$ more robust against a set of adversarial attacks $A_{attack}$ so that $T_{m}$ does not change its prediction on the set of adversarial samples $A_{test}$. We have limited access to the dataset $D_{train}$ due to data privacy concerns. Our objective is to improve the performance of $T_{m}$ on $A_{test}$ without significantly impacting its performance on $D_{test}$. As illustrated in Fig.~\ref{fig:overview}, we add a detection block before feeding the input to model $T_{m}$. We then pass all samples through a correction module to reduce adversarial contamination. The correction module is designed such that it does not denoise the clean images while it denoises adversarial images highly to reduce adversarial contamination. The corrected samples are then fed to pre-trained model $T_{m}$ to get the predictions. Next, we explain in detail our proposed detection and correction modules.
\begin{figure*}[htp]
\centering
\centerline{\includegraphics[width=0.99\textwidth]{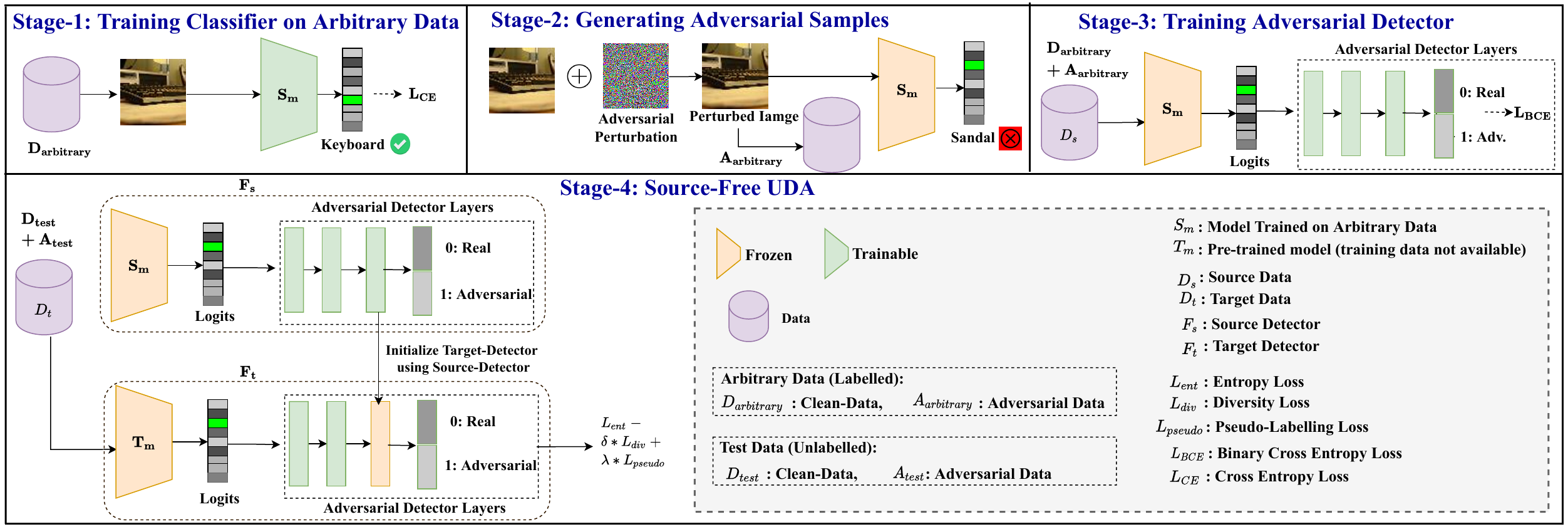}}
\caption{Our proposed detection module depicts different operations in each of the stages. We formulate adversarial detection as a source-free UDA problem where we adapt the source-detector trained on arbitrary data to the target detector using unlabelled clean and adversarial test samples.
}
\label{fig:detection}
\end{figure*}
\subsection{Detection Module}
\label{subsec:detection}
The core concept is that if we have access to an adversarial detector that can distinguish samples from an arbitrary dataset into either clean or adversarial, this binary classifier can be considered as a source classifier. The adversarial detector for the target model $T_{m}$ can be considered as a target classifier with the target data being the collection of unlabelled test samples and their corresponding adversarial samples. Therefore, as described in the introduction to UDA, we can formulate the adversarial detection as a UDA problem where:\\
$F_{s} \leftarrow$ Model $S_{m}$ appended with detection layers $L_{advdet}$\\
$D_{s} \leftarrow$ Mix of $D_{arbitrary}$ (clean) and $A_{arbitrary}$ (adversarial)\\
$D_{t} \leftarrow$ Mix of $D_{test}$ (clean) and $A_{test}$ (adversarial)\\
$F_{t} \leftarrow$ Model $T_{m}$ appended with detection layers $L_{advdet}$\\
\par The objective of our detection module is to use UDA techniques to obtain a model $F_{t}$ that can classify samples from $D_{t}$ as either clean or adversarial, by reducing the domain shift between $D_{s}$ and $D_{t}$. The detection module comprises four stages, as shown in detail in Figure~\ref{fig:detection}. Each stage performs a specific task, which is described below: \\
\textbf{Stage-1}: The first stage is to train a model $S_{m}$ using labeled data $D_{arbitrary}$ by minimizing the cross-entropy loss: $\min \sum_{i=1}^{M} L_{ce}(S_{m}(x_{iA}), y_{iA})$.\\
\textbf{Stage-2}: The second stage is to generate a set of adversarial samples ($A_{arbitrary}$) by using any adversarial attack $A_{j}$ from a set of adversarial attacks ($A_{attack}$), such that it deceives the trained network $S_{m}$.\\
\textbf{Stage-3}: In the third stage, the adversarial detection layers ($L_{advdet}$) are trained using the logits of network $S_{m}$ as input and the soft score for the two classes (adversarial and clean) as output. The architecture of these layers, similar to \cite{jonathan2019introspect}, consists of three fully connected layers containing 128 neurons and ReLU activations (except for last layer). The first two layers are followed by a dropout layer with a 25\% dropout rate. Additionally, the second (after-dropout) and third layers are followed by a Batch Normalization and Weight Normalization layer, respectively. The training is done using binary cross entropy loss $L_{BCE}$ on data $D_{s}$. \\
\textbf{Stage-4}:  The fourth stage is to perform UDA with the source network being $F_{s}$ and the target network being $F_{t}$, where the source data is $D_{s}$ and target data is $D_{t}$. If we remove the dependency on dataset $D_{s}$, then it also facilitates the use of any off-the-shelf pre-trained detector classifier in our framework and stages $1$ to $3$ can be skipped in that case. Therefore, to make our framework completely data-free and reduce its dependency on even arbitrary data, we adopt source-free UDA. To the best of our knowledge, the UDA setup for adversarial detection has not been previously explored, even for data-dependent approaches.
\par The target network $F_{t}$ is trained on an unlabelled dataset $D_{t}$ with the network $T_{m}$ being frozen and the layers of $L_{advdet}$ of $F_{t}$ remaining trainable, except for the last classification layer. These layers are initialized with the weights of $L_{advdet}$ of $F_{s}$. As per the methods described in ~\cite{liang2020we}, three different losses are used during training. The entropy loss ($L_{ent}$) is minimized to ensure that the network $F_{t}$ is confident in its predictions for each sample, making strong predictions for one of the classes (adversarial or clean). However, this could result in a degenerate solution where only one class is predicted. To avoid this outcome, the diversity loss ($L_{div}$) is maximized, which ensures high entropy across all samples, that is enforcing the mean of network predictions to be close to a uniform distribution. Even if the ($L_{ent}-L_{div}$) is minimized, there is still a possibility of unlabelled samples in $D_{t}$ getting assigned a wrong label. To address this, the pseudo-labels are estimated through self-supervision~\cite{caron2018deep,liang2020we}. The initial centroid for each class is calculated by taking the weighted mean of all samples, with the weight of each sample being its predicted score for that class. The cosine similarity between each sample and the centroids is then calculated, and the label of the nearest centroid is used as the initial pseudo-label. The initial centroids and pseudo-labels are then updated using a method similar to K-means. The overall loss $L = (L_{ent}-\delta L_{div}+ \lambda L_{pseudo})$ is minimized, where $\delta$ and $\lambda$ are hyperparameters, and $L_{pseudo}$ is the cross-entropy loss on $D_{t}$ using the assigned pseudo-labels as ground truth labels.
\subsection{Correction Module}
\begin{table*}[hbp]
\centering
\caption{Ablation on radius selection. Comparison of our proposed technique against a random baseline (R. B.) in terms of correction adversarial accuracy (Co.A). Our method shows significant improvement over the random baseline. The random baseline involved selecting the radius randomly and the results are reported as the mean over five trials.}
\label{tab:ablation}
\scalebox{0.8}{
\begin{tabular}{ccllllll}
\hline
\multicolumn{1}{l}{\multirow[b]{2}{*}{\textbf{Dataset}}} &
  \multicolumn{1}{l}{\multirow[b]{2}{*}{\textbf{Model}}} &
  \multicolumn{2}{c}{\textbf{PGD}} &
  \multicolumn{2}{c}{\textbf{IFGSM}} &
  \multicolumn{2}{c}{\textbf{Auto-Attack}} \\ \cmidrule{3-8} 
\multicolumn{1}{l}{} &
  \multicolumn{1}{l}{} &
  \multicolumn{1}{c}{\textbf{R. B.}} &
  \multicolumn{1}{c}{\textbf{Ours}} &
  \multicolumn{1}{c}{\textbf{R. B.}} &
  \multicolumn{1}{c}{\textbf{Ours}} &
  \multicolumn{1}{c}{\textbf{R. B.}} &
  \multicolumn{1}{c}{\textbf{Ours}} \\ \hline \hline
\multirow{3}{*}{CIFAR} & VGG-16   & 28.29 & 41.04 & 36.22 & 38.37 & 28.51  & 40.06 \\ 
                       & Resnet18 & 23.15 & 39.39 & 28.12 & 38.49 & 24.82 & 40.25 \\ 
                       & Resnet34 & 23.67 & 41.71 & 25.39 & 40.62 & 21.64 & 42.4 \\ \hline
\multirow{3}{*}{FMNIST} & VGG-16   & 10.95 & 33.09 & 13.84 & 33.83 & 9.64 & 37.99 \\ 
                       & Resnet18 & 8.55 & 32.22 & 12.87 & 32.38 & 9.50 & 35.8 \\ 
                       & Resnet34 & 9.82 & 33.23  & 10.69 & 33.73  & 9.47 & 35.82 \\ \hline
\end{tabular}}
\end{table*}

The input images are then processed by the correction module, which aims to reduce the contamination caused by adversarial attacks on the images. The model's prediction can be easily altered by adversarial samples, and it has been observed that the model's predictions are highly correlated with the high-frequency components (HFC) of the input image~\cite{wang2020high}. To counter this, we aim to eliminate the contaminated HFC from the image. This is also in line with human cognition, as humans tend to make decisions based on the low-frequency components (LFC) and ignore the HFCs~\cite{wang2020towards}. As a result, the HFCs are ignored at a certain radius $r$. For the $i^{th}$ image $d_{i}$ (refer to the Fourier Transform operations in the Preliminaries Section~\ref{sec:prelims}), we have:
\begin{equation}\label{eq3}
\begin{gathered}
f_i = F(d_{i})\\
\text{At radius $r$}: fl_{ir} = LFC(f_{i}, r) ,  fh_{ir} = HFC(f_{i}, r)  \\
dl_{ir} = F^{-1}(fl_{ir})
\end{gathered}
\end{equation}

The LFCs obtained by selecting a random radius $r$ may not always yield the desired results, as per the results shown in Table~\ref{tab:ablation}. LFCs at high radius provide better discriminability (Disc) but also result in higher adversarial contamination (AdvCont). On the other hand, LFCs at small radius results in lower AdvCont but also lower Disc. Therefore, a careful selection of the appropriate radius $r^{*}$ is crucial to achieving a balance between Disc and AdvCont, that is, to have low AdvCont but still maintain a high level of Disc. Thus, the corrected $i^{th}$ image in the spatial domain at radius $r^{*}$ is presented as:
\begin{equation}\label{eq4}
\begin{aligned}
d_{ic} =  dl_{ir^{*}} = F^{-1}(fl_{ir^{*}})  
\end{aligned}
\end{equation}

\par Determining the value of the radius $r^{*}$ can be a difficult task, particularly when there is no prior information about the training data. Additionally, the optimal value of $r^{*}$ may vary for each individual sample. In order to address this issue, we propose a correction method that calculates $r^{*}$ for each sample, and returns the corrected sample, as outlined in Algorithm~\ref{algo_corrections}. We first establish the minimum and maximum radius, $r_{min}$ and $r_{max}$, as the initial parameters (Line~\ref{initial}). As discussed, there are two key factors to consider when selecting the radius: Disc and AdvCont. To evaluate these factors, we measure the Disc and AdvCont present in LFC at different radii between $r=2$ and $r_{max}$, with a step size of 2. We use SSIM~\cite{wang2004image} as a perceptual metric to quantify the Disc for each sample. Specifically, the Disc score for the $i^{th}$ sample (Line~\ref{disc_score}) is calculated as:
\begin{equation}\label{eq5}
\begin{aligned}
Disc_{score}(d_{i}, dl_{ir}) = SSIM(d_{i}, dl_{ir})
\end{aligned}
\end{equation}
The SSIM score should be high to ensure good discriminability. To standardize the scores, the SSIM score is normalized between $0$ to $1$. While increasing the radius can improve the $Disc_{score}$, it also allows for more adversarial perturbations (often found in HFC regions) to pass through. As such, it is also important to evaluate the AdvCont, which measures how much perturbations have infiltrated the LFC in relation to image $d_{i}$. To address this, we calculate the label-change-rate (LCR) at each radius. The core idea behind our method is that if a significant number of perturbations have passed through at a certain radius $r$, the adversarial component would become the dominant factor, causing the low-pass spatial sample $dl_{ir}$ to have the same prediction as $d_{i}$. To obtain a more accurate estimate, we repeatedly compute the low-pass predictions after enabling dropout. 

\begin{algorithm}[hb]
\small
 \caption{Correction of image samples ($D_{t}$) by determining the best radius $r^{*}$}
 \label{algo_corrections}
\begin{algorithmic}[1]
 \renewcommand{\algorithmicrequire}{\textbf{Input:}}
 \renewcommand{\algorithmicensure}{\textbf{Output:}}
 \Require Pre-trained model $T_{m}$, $d_{i}$: $i^{th}$ image sample from $D_{t}$, $F_{t}$: Target detector, $w$: Maximum frequency in image after Fourier transform.
 \Ensure $d_{ic}$ : corrected $i^{th}$ sample
 \label{detection}
 \State Obtain softmax output of target detector on $d_{i}$ sample: $p_{i}^{d} \leftarrow F_{t}(d_{i})$
 \State Probability of image sample as clean $p_{i}^{cd} \leftarrow p_{i}^{d}[0]$ 
 \State Obtain prediction on the $i^{th}$ sample: $pred \leftarrow label(T_{m}(d_{i}))$
 \State Initialize: $r_{max}\leftarrow\frac{w}{2}$, $r_{min}\leftarrow w\cdot p_{i}^{cd}$ , $count\leftarrow10$ \label{initial}
 \vspace{-0.12in}
 \State Set dropout in training mode for $T_{m}$ \label{drop}
 \For{$r=2;\ r \le r_{max};\ r = r + 2$:}
 \State Obtain LFC $dl_{ir}$ for $i^{th}$ sample $d_{i}$ using eq.~\ref{eq3}
 \State Compute discriminability score for $dl_{ir}$: $Disc\_dl_{ir} \leftarrow Disc_{score}(d_{i}, dl_{ir})$ using eq.~\ref{eq5} \label{disc_score}
 \State Initialize the label change rate: $lcr_{r}=0$ \label{initial_lcr}
 \For{$k=1:count$}
 \State $pred_{r} = label(T_{m}(dl_{ir}))$
 \If{$pred_{r} \ne pred$}
 \State $lcr_{r} = lcr_{r} + 1$  \label{lcr_end}
 \EndIf
 \EndFor
 \State $AdvCont\_dl_{ir} \leftarrow AdvCont_{score}(dl_{ir})= (count-lcr_{r}) / count$ \label{normalize}
 \If{($Disc\_dl_{ir} - AdvCont\_dl_{ir}) > 0$}  \label{max_r_start}
 \State $r' = r$
 \Else
 \State break  \label{value_change}
 \EndIf

\EndFor

 \State $r^{*} = max(r_{min} , r')$  \label{max_r_end}
 \State Obtain LFC at best radius $r^{*}$: $\newline$ $d_{ic} = dl_{ir^{*}} = F^{-1}(LFC(F(d_{i}), r^{*}))$ \label{best_r}

\end{algorithmic} 
 \end{algorithm}
By enabling dropout, the decision boundary is slightly perturbed. If adversarial noise has begun to dominate, the shift in the decision boundary will have little to no effect on the model's prediction. To quantify this effect, the LCR (at a specific radius $r$) records the number of times the LFC prediction differs from the original adversarial sample (Lines~\ref{initial_lcr}-~\ref{lcr_end}). In other words, a higher LCR indicates low adversarial contamination and vice versa.

Line~\ref{normalize} normalizes $AdvCont_{score}$ between $0$ to $1$. This allows us to directly compare the adversarial and discriminability scores. The optimal radius for $d_{i}$ is the maximum radius at which $Disc_{score}$ is greater than $AdvCont_{score}$ (Lines~\ref{max_r_start}-\ref{max_r_end}).Finally, we set the value of $r^{*}$ to the maximum value between $r_{min}$ and $r'$(Line~\ref{value_change}). The corrected $i^{th}$ sample i.e. $d_{ic}$ (Line\ref{best_r}) is then passed to the model $T_{m}$ for predictions.

\subsection{Soft Vs Hard Detection}\label{softvshard}

In our previous framework DAD, we passed input images to the detector, which detected images as being clean or adversarial. If the image is clean, it is directly fed to the target model. On the other hand, if it is detected as adversarial, it is first corrected by passing through the correction module and then passed to the target model (termed as 'Hard detection’). However, in cases when the detector is not confident about the prediction of an image, it is likely to misclassify the image. This leads to a huge drop in overall performance. To avoid such cases, in this paper we introduce a ‘Soft detection’ scheme. Unlike the ‘Hard detection’ mechanism, which utilized the binary output vector from the detector, our novel ‘Soft detection’ method takes into account the softmax output from the detector. In this approach, we pass all images from the dataset $D_{t}$ to a correction module. For each image, the probability of the image being clean, $p_{i}^{cd}$, is obtained from the detector. The $r_{min}$ is then set as the product maximum frequency in image $w$ (It is same as image size) and $p_{i}^{cd}$. In the end, we set the value of $r^*$ to the maximum between radius computed by correction algorithm $r'$ and $r_{min}$. In next few paragraphs we explain how our Soft Detection method improves the clean accuracy of the target model at the cost of small drop in adversarial accuracy compared to Hard Detection.

The correction algorithm in our conference paper DAD~\cite{Nayak2022DADDA} essentially identifies the maximum radius, $r^*=r$, at which the model's prediction($pred_{{r}^{*}}$), differs from the original prediction ($pred$). Therefore, if the original prediction, $pred$, is incorrect (i.e, it differs from the ground truth), the prediction for the denoised image,$pred_{{r}^{*}}$, obtained using the maximum radius $r^{*}$ is more likely to be correct. On the other hand, if $pred$ is correct, then $pred_{{r}^{*}}$ will be incorrect. In DAD, we used a hard detection approach, where the detector predicts whether an image is clean or adversarial, and the correction module is used to denoise all adversarial images. However, if the detector has a high percentage of false positives, a large number of clean images for which $pred$ is correct, will be denoised, leading to an incorrect prediction $pred_{{r}^{*}}$, This eventually causes a significant decrease in the model's combined clean accuracy.

In DAD++, since we are passing all images to the correction module, we need the correction algorithm to ensure that for clean images, $pred_{{r}^{*}}$ is the same as $pred$ and, for adversarial images, $pred_{{r}^{*}}$ should be different from $pred$. In other words, for clean images, we want very little or no denoising of the image, while for adversarial images, we want the image to be highly denoised to remove the adversarial contamination. In order to demonstrate that this is achieved by DAD++, we consider the following scenarios.

Before that, in DAD++, correction algorithm essentially outputs maximum radius $r'$ at which the model's prediction($pred_r'$), differs from the original prediction($pred$).  if $r' < r_{min}$, then the prediction at $r_{min}$ is likey to be same as the original prediction of the model. Conversely, if $r' > r_{min}$, then the prediction at $r_{min}$ is different from the original prediction.

In the first scenario, if the detector accurately predicts a clean image with high confidence (i.e. $p_{i}^{cd} \approx1$), the value of $r_{min}$ will be high. Consequently, for most images, $r' < r_{min}$ and $r^* = r_{min}$, which implies $pred_{{r}^{*}}$ = $pred$. For example, if the image size is 32 and $p_{i}^{cd} = 0.9$, $r_{min}$ will be 28. We have observed that at such a high radius, the model's prediction does not change after denoising (i.e. $pred_{{r}^{*}} = pred$).

In the second scenario, if the detector accurately predicts an adversarial image with high confidence (i.e. $p_{i}^{cd} \approx 0$), the value of $r_{min}$ will be close to zero. Therefore, for most images, $r' > r_{min}$ and  $r^* > r_{min}$, which implies $pred_{{r}^{*}} \neq pred$. For example, if $p_{i}^{cd} = 0.1$, $r_{min}$ will be $3$. For most images, $r'$ would likely to be greater than $3$.

In the third scenario, if the detector incorrectly predicts an image as adversarial with low confidence, the value of $r_{min}$ will be neither very low nor very high. As a result, some clean images will have $r' < r_{min}$, leading to $pred_{{r}^{*}} = pred$. In contrast, the hard detection method would always output $r^{*}$ such that, $pred_{{r}^{*}} \neq pred$ for these samples, the resulting drop in clean accuracy. However, by using soft detection, we can obtain some improvement in clean accuracy for this scenario.

In the fourth scenario, if the detector correctly predicts an image as adversarial with low confidence, the value of $r_{min}$ will be neither very low nor very high. Consequently, some adversarial images will have $r' < r_{min}$, leading to $pred_{{r}^{*}} = pred$. In the hard detection method, we would always get $pred_{{r}^{*}} \neq pred$ for these samples. However, with soft detection we get, $pred_{{r}^{*}} = pred$ for some images, which results in a decrease in the adversarial accuracy of the model (since $pred_{{r}^{*}} \neq pred$ is desired for adversarial images).

Based on the experiments conducted in Table~\ref{soft_vs_hard}, we observed that the soft detection method provides higher clean accuracy than the hard detection method, although its adversarial accuracy is lower. However, the improvement in clean accuracy is much more significant than the decrease in adversarial accuracy.

\section{Experimental evaluation and applications of DAD++}
\subsection{Soft vs Hard Detection}

As discussed in section~\ref{softvshard}, we propose a soft detection technique in this paper to overcome the drawbacks of hard detection. In this section, we demonstrate the superiority of soft detection over hard detection using the results reported in Table~\ref{soft_vs_hard}. We conduct our experiments on three target datasets, i.e., CIFAR10~\cite{krizhevsky2009learning}, Tiny Imagenet~\cite{le2015tiny}, and CUB~\cite{WahCUB_200_2011}. For CIFAR10, we take FMNIST~\cite{xiao2017fashion} as the arbitrary dataset, while  for Tiny ImageNet and CUB, we take CIFAR10. For all the datasets, we take Resnet18 as both the target and source model.

Across all the datasets, we observe that the clean accuracy of the target model is always higher with soft detection than with that hard detection. However, the adversarial accuracy slightly decreases with soft detection for the CUB dataset across attacks. For the CIFAR10 dataset, we observed that soft detection leads to better adversarial accuracy compared to hard detection when using the auto attack. The same phenomenon was observed for the Tiny ImageNet dataset.

\begin{table*}[t]
\caption{Notations - Hard Det.: Combined accuracy of target model with hard detection method, Soft Det.: Combined Accuracy of target model with soft detection method. Comparison of the combined accuracy of the target model using hard detection and soft detection methods. The results show that the soft detection method results in high clean accuracy with a small decrease in adversarial accuracy compared to the hard detection method.}
\label{soft_vs_hard}
\scalebox{0.77}{
\begin{tabular}{ccccccccc}
\hline
\multirow[b]{4}{*}{\textbf{\begin{tabular}[c]{@{}c@{}}Target\\ Dataset\end{tabular}}} &
\multirow[b]{4}{*}{\textbf{\begin{tabular}[c]{@{}c@{}}Target\\ Model\end{tabular}}} &
\multirow[b]{4}{*}{\textbf{\begin{tabular}[c]{@{}c@{}}Arbitrary\\ Dataset\end{tabular}}} &
\multicolumn{6}{c}{\textbf{Accuracy}} \\ \cmidrule{4-6}\cmidrule{7-9}
& & & 
\multicolumn{3}{c}{\textbf{PGD}} & 
\multicolumn{3}{c}{\textbf{Auto-Attack}} \\ \cmidrule{4-6}\cmidrule{7-9} 
& & & 
\multirow{2}{*}{\textbf{\begin{tabular}[c]{@{}c@{}}Baseline\end{tabular}}} &
\multirow{2}{*}{\textbf{\begin{tabular}[c]{@{}c@{}}DAD \\ (Hard Det)\end{tabular}}} &
\multirow{2}{*}{\textbf{\begin{tabular}[c]{@{}c@{}}DAD++ \\ (Soft Det)\end{tabular}}} &
\multirow{2}{*}{\textbf{\begin{tabular}[c]{@{}c@{}}Baseline\end{tabular}}} &
\multirow{2}{*}{\textbf{\begin{tabular}[c]{@{}c@{}}DAD \\ (Hard Det)\end{tabular}}} &
\multirow{2}{*}{\textbf{\begin{tabular}[c]{@{}c@{}}DAD++ \\ (Soft Det)\end{tabular}}}  \\
\vspace{2\baselineskip}\\ \hline \hline
CIFAR10 & Resnet18 & FMNIST & 93.36 / 0 & 89.55 / 35.34 & 91.42 / 33.55 & 93.36 / 0 & 92.59 / 32.58 & 92.94 / 33.57  \\ \hline
Tiny Imagenet & Resnet34 & CIFAR10 & 70.34 / 0 & 65.67 / 32.33 & 67.98 / 32.58 & 70.34 / 0 & 61.61 / 35.68 & 66.97 / 36.46  \\ \hline
CUB & Resnet34 & CIFAR10 & 75.33 / 0 & 63.05 / 30.60 & 70.97 / 28.93 & 75.33 / 0.0 & 67.16 / 33.41 & 73.30 / 32.81 \\ \hline

\end{tabular}}
\end{table*}

In addition to the results presented here, we compared the performance of soft and hard detection for all experiments conducted in sections~\ref{sl} to~\ref{uda}. We consistently observe that soft detection yields higher clean accuracy than hard detection, even though there is a slight decrease in adversarial accuracy for some datasets. For real-world datasets such as CUB, Tiny ImageNet, Office-31~\cite{saenko2010adapting}, and Office Home~\cite{venkateswara2017deep}, we observed that the improvement in clean accuracy is much more significant than the decrease in adversarial accuracy. In the rest of the paper, results are reported using the soft detection method only.
\subsection{Supervised Classification}\label{sl}
\begin{table*}[htp]
\caption{Notations- Baseline: Accuracy of target model without adversarial defense, TD.A:Target detector accuracy, Co.A:Correction accuracy, Cb.A:Combined accuracy. Performance of DAD++ defense on target model trained using supervised training. The results show that DAD++ is able to provide a strong defense against PGD and Auto Attack to the target model}
\label{vanilla}
\scalebox{0.77}{
\begin{tabular}{cccccccccc}
\hline
\multirow[b]{3}{*}{\textbf{\begin{tabular}[c]{@{}c@{}}Target\\ Dataset\end{tabular}}} &
\multirow[b]{3}{*}{\textbf{\begin{tabular}[c]{@{}c@{}}Target\\ Model\end{tabular}}} &
\multirow[b]{3}{*}{\textbf{\begin{tabular}[c]{@{}c@{}}Arbitrary\\ Dataset\end{tabular}}} &
\multirow[b]{3}{*}{\textbf{\begin{tabular}[c]{@{}c@{}}Arbitrary\\ Model\end{tabular}}} &
\multicolumn{6}{c}{\textbf{Accuracy}} \\ \cmidrule{5-10}
& & & & 
\multicolumn{3}{c}{\textbf{PGD}} & 
\multicolumn{3}{c}{\textbf{Auto-Attack}} \\ \cmidrule{5-10} 
& & & &
\textbf{TD.A} &
\textbf{Co.A} &
\textbf{Cb.A} &
\textbf{TD.A} &
\textbf{Co.A} &
\textbf{Cb.A} \\ \hline \hline 
\multirow{4}{*}{CIFAR10} &
\multirow{4}{*}{Resnet18} &
\multicolumn{2}{c}{\textbf{Baseline}} & & &
93.36 / 0 & & & 
93.36 / 0 \\ 
\cmidrule{3-10}
& & FMNIST & Resnet18 & 91.62 / 91.33 & 41.06 & 91.42 / 33.55 & 98.24 / 83.46 & 39.81 & 92.94 / 33.57 \\
\cmidrule{3-10}
& & MNIST & Resnet18 & 92.07 / 88.80 & 41.06 & 91.08 / 32.43 & 96.61 / 90.93 & 39.81 & 92.19 / 36.83 \\
\cmidrule{3-10}
& & SVHN & WRN & 99.72 / 93.88 & 41.06 & 93.23 / 36.88 & 99.80 / 92.05 & 39.81 & 93.31 / 36.33 \\
\hline
\multirow{4}{*}{\begin{tabular}[c]{@{}c@{}}Tiny\\ Imagenet\end{tabular}} &
\multirow{4}{*}{Resnet34} &
\multicolumn{2}{c}{\textbf{Baseline}} & & &
70.34 / 0 & & & 
70.34 / 0 \\ 
\cmidrule{3-10}
& & FMNIST & Resnet18 & 86.03 / 86.75 & 34.12 & 63.20 / 29.46 & 71.68 / 72.15 & 39.9 & 55.60 / 35.33  \\ \cmidrule{3-10}
& & CIFAR10 & Resnet18 & 93.40 / 94.87 & 34.12 & 67.98 / 32.58 & 88.30 / 72.00 & 39.9 & 66.97 / 36.46  \\ \cmidrule{3-10}
& & SVHN & WRN & 99.98 / 99.22 & 34.12 & 70.34 / 33.69 & 98.04 / 70.60 & 39.9 & 70.09 / 36.75 \\ \hline
\multirow{4}{*}{CUB} &
\multirow{4}{*}{Resnet34} &
\multicolumn{2}{c}{\textbf{Baseline}} & & &
75.33 / 0 & & & 
75.33 / 0 \\ 
\cmidrule{3-10}
& & FMNIST & Resnet18 & 75.87 / 93.25 & 33.18 & 65.34 / 28.81 & 75.9 / 76.09 & 39.55 & 66.98 / 32.91 \\ \cmidrule{3-10}
& & CIFAR10 & Resnet18 & 84.05 / 94.11 & 33.18 & 70.97 / 28.93 & 89.85 / 74.24 & 39.55 & 73.30 / 32.81 \\ \cmidrule{3-10}
& & SVHN & WRN & 98.96 / 96.87 & 33.18 & 75.32 / 26.32 & 98.03 / 74.45 & 39.55 & 75.27 / 31.17

 \\ \hline
\end{tabular}}
\end{table*}

In this subsection, we analyze the performance of our proposed methodology, DAD++, in a supervised classification framework where a target model is trained using labeled data in a supervised learning method. We aim to provide a test-time defense to the pre-trained model using our methodology without any re-training. To check the effectiveness of our defense, we conduct experiments on three target datasets: CIFAR10, CUB, and Tiny Imagenet. The target model for CIFAR10 is Resnet18; and Resnet34 for CUB and Tiny Imagenet datasets. An arbitrary model is Resnet18 and Wideresnet-16-1. For training arbitrary models, we select the arbitrary dataset from FMNIST, MNIST~\cite{lecun1998gradient}, SVHN~\cite{SVHN}, and CIFAR10. The arbitrary dataset for a particular target dataset is selected in a manner such that the distribution of the target and arbitrary dataset is different. Results are presented in Table~\ref{vanilla}. For the CIFAR10 dataset, we achieved the best results with SVHN and Wideresnet (WRN) as the arbitrary dataset and the arbitrary model. We observe a small drop of 0.13\% in the clean accuracy, while the adversarial accuracy increases by 36.88\% compared to the baseline against the PGD attack. 

\begin{table*}[hbp]
\centering
\caption{Comparison of DAD++ with state-of-the-art adversarial defense methods. The target model is Resnet50 and Target dataset is CIFAR10. Our method does not require training data to train the defense network, and it also provides defense to pre-trained target models without retraining.
}
\label{comparison}
\scalebox{0.8}{
\begin{tabular}{ccccc}
\hline

\multirow{3}{*}{\textbf{\begin{tabular}[c]{@{}c@{}}Defense\\Method\end{tabular}}} & 
\multirow{3}{*}{\textbf{\begin{tabular}[c]{@{}c@{}}Data\\free\end{tabular}}} & 
\multirow{3}{*}{\textbf{\begin{tabular}[c]{@{}c@{}}Defends\\pre-trained\\ model\end{tabular}}} & 
\multirow{3}{*}{\textbf{\begin{tabular}[c]{@{}c@{}}Clean\\Accuracy\end{tabular}}} & 
\multirow{3}{*}{\textbf{\begin{tabular}[c]{@{}c@{}}PGD\\Accuracy\end{tabular}}} \\ 
& & & & \\ 
& & & & \\ \hline
Baseline & - & - & 93 & 0\\ \hline
Adv. Training ~\cite{Madry2017TowardsDL} & \xmark & \xmark & 83.53 & 46.07 \\ \hline
NRP~\cite{Naseer2020ASA} & \xmark & \cmark & 90.9 & 35.9  \\ \hline
ComDefend ~\cite{Jia2018ComDefendAE} & \xmark & \cmark & 89.5 & 50.1 \\ \hline
SR ~\cite{Mustafa2019ImageSA} & \cmark & \cmark & 45.2 & 30.3 \\ \hline
DIPDefend~\cite{Dai2020DIPDefendDI} & \cmark & \cmark & 83.2 & 73.6 \\ \hline
DIDIP ~\cite{Ding2021DelvingID} & \cmark &  \cmark &  85.6 & 83.1  \\ \hline
DAD ~\cite{Nayak2022DADDA} & \cmark & \cmark & 83.11 & 50.24 \\ \hline
DAD++ (ours) & \cmark & \cmark & 91.5 & 48.51 \\ \hline

\end{tabular}
}
\end{table*}

\par For Tiny Imagenet and CUB, we observe a similar trend where CIFAR10 and SVHN as the arbitrary dataset, produce better results than FMNIST. Overall, the combination of SVHN and WRN as the arbitrary dataset and model ensures the best results for both CUB and Tiny Imagenet. For the Tiny Imagenet dataset, compared to the baseline, we notice no drop in clean accuracy while the adversarial accuracy increases to 33.69\% against the PGD attack. Against the Auto attack, the clean accuracy drops by only 0.25\% while the adversarial accuracy increases to 36.75\%. For CUB, we observe a small drop in clean accuracy while the adversarial accuracy increases to 26.32\% and 31.17\% against PGD and Auto attacks, respectively.

\par Table~\ref{comparison} presents the results of state-of-the-art methods for adversarial defense against PGD attack on Resnet50 model trained on Cifar10 dataset. Our method yields better adversarial accuracy than SR~\cite{Mustafa2019ImageSA}, NRP~\cite{Naseer2020ASA}, and Adversarial training~\cite{Madry2017TowardsDL}. However, DIDIP~\cite{Ding2021DelvingID}and DIPDefend~\cite{Dai2020DIPDefendDI} achieve superior adversarial accuracy compared to our approach. It is important to note that these two methods employ deep image prior, which updates the denoising network's parameters using backpropagation for every single image. Therefore DIP based methods cannot process images in batches. Furthermore, DIDIP requires 1000 backpropagation iterations to denoise the image, and DIPDefend takes around 2000 iterations. Because of this, DIP-based methods are slow and unsuitable for real-world applications.  We computed the time required for the execution of DIDIP and DIPDefend on 200 images from the CIFAR10 dataset (100 clean images and 100 adversarial images) and present the results in Table~\ref{timecomparison}. We used the hyperparameters reported in the respective papers. For these images, DIDIP takes $\approx$38 minutes to process, whereas DIPDefend takes more than one hour. In comparison, our method takes very little time to process( only 10 seconds). The same trend is observed when the number of images in the test sets increases. When the number of images is increased to 800 in the test set, DIDIP, and DipDefend, required approximately 157 minutes and 251 minutes, while DAD++ can process the same number of images in just 1.7 minutes. It could be concluded from the observation that despite their superior adversarial accuracy, DIP-based methods may not be practical for real-world adversarial defense tasks due to their slow processing times. On the other hand, our proposed method, DAD++, is much more feasible to use in real-world applications with a minimal amount of inference time.

\begin{table}[t]
\centering
\caption{Comparison of Test Runtime of DIDIP, DIPDefend and DAD++ 
}
\label{timecomparison}
\begin{tabular}{cccc}
\hline
\multirow{3}{*}{\textbf{\begin{tabular}[c]{@{}c@{}}Number of \\ Samples\end{tabular}}} & 
\multicolumn{3}{c}{\textbf{Runtime ( In minutes)}}\\ \cmidrule{2-4}
& \multirow{2}{*}{\textbf{\begin{tabular}[c]{@{}c@{}}DIDIP \\ ~\cite{Ding2021DelvingID}\end{tabular}}} & \multirow{2}{*}{\textbf{\begin{tabular}[c]{@{}c@{}}DIPDefend \\ ~\cite{Dai2020DIPDefendDI}\end{tabular}}} & \multirow{2}{*}{\textbf{\begin{tabular}[c]{@{}c@{}}DAD++ \\ (Ours)\end{tabular}}} \\
\vspace{2\baselineskip}\\ \hline
200 & 38.14 & 64.01 & 0.7 \\ \hline
400 & 80.22 & 126.78 & 1.2 \\ \hline
600 & 120.01 & 189.28 & 1.5 \\ \hline
800 & 157.04 & 251.31 & 1.7 \\ \hline
\end{tabular}
\end{table}


\begin{table*}[hbp]
\caption{Performance of the (DAD++) against multiple attacks on the target model. The results indicate that DAD++ provides good defense against multiple attacks, even when the adversarial test contains images created using different attack methods. Notations- TD.A:Target detector accuracy, Co.A:Correction accuracy, Cb.A:Combined accuracy.}
\label{combined_attack}
\scalebox{0.8}{
\begin{tabular}{ccccccccc}
\hline
\multirow[b]{3}{*}{\textbf{\begin{tabular}[c]{@{}c@{}}Target\\ Dataset\end{tabular}}} &
\multirow[b]{3}{*}{\textbf{\begin{tabular}[c]{@{}c@{}}Target\\ Model\end{tabular}}} &
\multirow[b]{3}{*}{\textbf{\begin{tabular}[c]{@{}c@{}}Arbitrary\\ Dataset\end{tabular}}} &
\multicolumn{6}{c}{\textbf{Accuracy}} \\ \cmidrule{4-9}
& & & 
\multicolumn{3}{c}{\textbf{PGD / Auto-Attack (0.5/0.5)}} & 
\multicolumn{3}{c}{\textbf{PGD / Auto-Attack / BIM (0.2/0.4/0.4)}} \\ \cmidrule{4-9} 
& & &
\textbf{TD.A} &
\textbf{Co.A} &
\textbf{Cb.A} &
\textbf{TD.A} &
\textbf{Co.A} &
\textbf{Cb.A} \\ \hline \hline 
\multirow{3}{*}{CIFAR10} &
\multirow{3}{*}{Resnet18} &
\multicolumn{1}{c}{\textbf{Baseline}} & & &
93.36 / 0 & & &
93.36 / 0 \\ 
\cmidrule{3-9}
& & FMNIST & 95.09 / 83.95 & 40.62 & 92.52 / 32.55 & 95.02 / 87.01 & 39.44 & 92.49 / 30.67  \\
\cmidrule{3-9}
& & MNIST & 91.15 / 89.02 & 40.62 & 90.13 / 34.33 & 95.88 / 83.50 & 39.44 & 91.68 / 30.65 \\
\hline
\multirow{3}{*}{CUB} &
\multirow{3}{*}{Resnet34} &
\multicolumn{1}{c}{\textbf{Baseline}} & & &
75.33 / 0 & & &
75.33 / 0 \\ 
\cmidrule{3-9}
& & FMNIST & 80.56 / 84.03 & 35.50 & 68.47 / 30.13 & 70.83 / 88.78 & 34.44 & 64.69 / 29.20  \\ \cmidrule{3-9}
& & CIFAR10 & 89.35 / 84.36 & 35.50 & 73.35 / 31.07 & 90.07 / 84.25 & 34.44 & 73.70 / 31.14 \\ \hline
\end{tabular}}
\end{table*}

We conclude with several key takeaways. Firstly, our proposed method DAD++ has been demonstrated to be effective for various datasets and model architectures. We have observed that DAD++ works well for toy datasets like CIFAR10, as well as providing robust defense for models trained on real-world datasets such as Tiny Imagenet and CUB. Secondly, our results suggest that using an arbitrary dataset containing RGB images like CIFAR10 or SVHN leads to better adaptation to the target detector than using simple grayscale image datasets like FMNIST and MNIST. Overall, our proposed method DAD++ effectively defends any pre-trained model in the absence of training data.
\subsection{Performance Against Multiple Attacks Combined}

In the previous section, we evaluated the performance of our DAD++ method on adversarial test sets created using a single type of attack (.i.e, either PGD or Auto Attack). In this section, we examine the performance of DAD++ in a more challenging scenario where the adversarial dataset contains images created using different types of attacks (PGD, Auto Attack, and BIM). For this, we consider two test scenarios. In the first case, we take our adversarial test set with 50\% images created using PGD attack and 50\% using Auto Attack. In the second case, we take the combination of PGD, BIM, and Auto Attack with their proportion of 20\%, 40\%, and 40\% in the adversarial test set. We report the results corresponding to the two cases in table~\ref{combined_attack}

For CIFAR10, we find that the clean accuracy drops by less than 1\% when an arbitrary dataset is the FMNIST dataset, and, the adversarial accuracy increases by more than 30\% compared to the baseline. Similarly, with the MNIST as an arbitrary dataset,  the clean accuracy decreases by $\approx$3-4\% with a gain of more than 30\% in adversarial accuracy against different attacks. 
 For the CUB dataset, we find that the clean accuracy of the model decreases by  $\approx$7-10\% when using FMNIST as an arbitrary dataset, while the clean accuracy decreases by only about 2\% when CIFAR10 is the arbitrary dataset. In both cases, the adversarial accuracy of the model increases by $\approx$30\%.

From the above observations, we conclude that the performance of our DAD++ defense is not affected if the adversarial test set contains images created using multiple attacks. Unlike the different adversarial training approaches, which are not able to perpetuate their efficacy against the multiple attacks in the test set, our method effectively combats such situations.

\subsection{Performance Against Imbalanced Quantity Of Clean and Adversarial Sample.}
\begin{table*}[hbp]
\centering
\caption{Performance of DAD++ against the imbalanced quantity of clean and adversarial sample. The experiments are done using two sets with a higher fraction of clean samples in one set and a higher fraction in the other. DAD++ shows consistent performance in these demanding setups. Notations- TD.A:Target detector accuracy, Co.A:Correction accuracy, Cb.A:Combined accuracy.}
\label{imbalanced_testset}
\scalebox{0.77}{
\begin{tabular}{cccccccccc}
\hline
\multirow[b]{3}{*}{\textbf{\begin{tabular}[c]{@{}c@{}}Target\\ Dataset\end{tabular}}} &
\multirow[b]{3}{*}{\textbf{\begin{tabular}[c]{@{}c@{}}Target\\ Model\end{tabular}}} &
\multirow[b]{3}{*}{\textbf{\begin{tabular}[c]{@{}c@{}}Clean:Adv \\ ratio\end{tabular}}} &
\multirow[b]{3}{*}{\textbf{\begin{tabular}[c]{@{}c@{}}Arbitrary\\ Dataset\end{tabular}}} &
\multicolumn{6}{c}{\textbf{Accuracy}} \\ \cmidrule{5-10}
& & & &
\multicolumn{3}{c}{\textbf{PGD}} & 
\multicolumn{3}{c}{\textbf{Auto-Attack}} \\ \cmidrule{5-10} 
& & & &
\textbf{TD.A} &
\textbf{Co.A} &
\textbf{Cb.A} &
\textbf{TD.A} &
\textbf{Co.A} &
\textbf{Cb.A} \\ \hline \hline
\multirow{4}{*}{CIFAR10} &
\multirow{4}{*}{Resnet18} &
\multirow{2}{*}{1:2} &
\multicolumn{1}{c}{\textbf{Baseline}} & & &
92.96 / 0.05 & & &
92.96 / 0.05 \\ 
\cmidrule{4-10}
& & & FMNIST & 99.73 / 86.94 & 41.66 & 92.87 / 31.85 & 99.04 / 90.37 & 39.95 & 92.95 / 36.15 \\ \cmidrule{3-10}
& & \multirow{2}{*}{2:1} &
\multicolumn{1}{c}{\textbf{Baseline}} & & &
93.06 / 0 & & &
93.06 / 0 \\ 
\cmidrule{4-10}
& & & FMNIST & 92.10 / 93.33 & 40.73 & 91.63 / 35.47 & 93.34 / 91.94 & 40.0 & 90.80 / 37.10\\ \cmidrule{1-10}
\multirow{4}{*}{CUB} &
\multirow{4}{*}{Resnet34} &
\multirow{2}{*}{1:2} &
\multicolumn{1}{c}{\textbf{Baseline}} & & &
72.15 / 0 & & &
72.15 / 0 \\ 
\cmidrule{4-10}
& & & CIFAR10 & 92.75 / 86.15 & 31.90 & 70.55 / 23.67 & 96.21 / 70.77 & 38.7 & 71.60 / 30.80  \\ \cmidrule{3-10}
& & \multirow{2}{*}{2:1} &
\multicolumn{1}{c}{\textbf{Baseline}} & & &
74.0 / 0 & & &
74.0 / 0 \\ 
\cmidrule{4-10}
& & & CIFAR10 & 97.00 / 84.37 & 31.35 & 71.15 / 29.10 & 77.14 / 79.19 & 38.6  & 67.37 / 34.00  \\ \cmidrule{1-10}
\end{tabular}}
\end{table*}
In this subsection, we test the sensitivity of our DAD++ approach to the imbalance proportion of clean and adversarial samples in the test set. 
For this, we consider two test scenarios with an imbalanced ratio of clean to adversarial samples and check the performance of our proposed DAD++ approach on it. In the first scenario, we take the number of adversarial samples two times more than the number of clean samples, and in the second scenario, we take clean samples two times more than the adversarial samples. We use the Resnet18 model as the arbitrary model. FMNIST is the arbitrary dataset for the CIFAR10 target dataset. For CUB, we use CIFAR10 as the arbitrary dataset. The results are reported in Table~\ref{imbalanced_testset}

For the CIFAR10 dataset, we observed that when the clean samples are more than adversarial samples, across attacks, the clean accuracy drops by only $\approx$1-2\%, and the adversarial accuracy increases by $\approx$36\%. When the adversarial samples are more, the drop in clean accuracy is minimal, with an increase of $\approx$32-36\% in adversarial accuracy when compared to the baseline.

For the CUB dataset, when adversarial samples are more than clean samples, compared to baseline, the clean accuracy decrease by $\approx$1-2\% across attacks. However, when the test set contains more clean samples, then we observe a more drop of $\approx$3-7\% in clean accuracy. However, in this case, the adversarial accuracy does not decrease too much compared to correction accuracy. 

In summary, our results demonstrate that the composition of the test set does have an impact on the detector accuracy, as reflected in the combined accuracy. However, it is important to note that this effect is not overly detrimental to our DAD++ defense. In particular, when the number of adversarial samples is greater than the number of clean samples, we obtain better detector accuracy than the other way around.

\subsection{Semi-Supervised Classification}\label{ssl}

\begin{table*}[t]
\centering
\caption{Performance comparison of the target model trained using the SSL-Flexmatch method with and without DAD++ defense. Target Model architecture is WRN and arbitrary model architecture is Resnet18 in these experiments. DAD++ shows significant improvement over the baseline setting while defending models trained using SSL-Flexmatch. Notations- TD.A:Target detector accuracy, Co.A:Correction accuracy, Cb.A:Combined accuracy.}
\label{SSL-Flexmatch}
\scalebox{0.8}{
\begin{tabular}{cccccccc}
\hline
\multirow[b]{3}{*}{\textbf{\begin{tabular}[c]{@{}c@{}}Target\\ Dataset\end{tabular}}} &
\multirow[b]{3}{*}{\textbf{\begin{tabular}[c]{@{}c@{}}Arbitrary\\ Dataset\end{tabular}}} &
\multicolumn{6}{c}{\textbf{Accuracy}} \\ \cmidrule{3-8}
& & 
\multicolumn{3}{c}{\textbf{PGD}} & 
\multicolumn{3}{c}{\textbf{Auto-Attack}} \\ \cmidrule{3-8} 
& &
\textbf{TD.A} &
\textbf{Co.A} &
\textbf{Cb.A} &
\textbf{TD.A} &
\textbf{Co.A} &
\textbf{Cb.A} \\ \hline \hline 
\multirow{4}{*}{CIFAR10} &
\multicolumn{1}{c}{\textbf{Baseline}} & & &
94.80 / 0 & & & 
94.80 / 0 \\ 
\cmidrule{2-8}
 & FMNIST &  86.82 / 83.64 & 46.65 & 91.85 / 35.26 & 85.75 / 89.06 & 46.12 & 91.24 / 41.02 \\
\cmidrule{2-8}
 & MNIST &  86.69 / 77.31 & 46.65 & 91.65 / 33.55 & 98.25 / 79.49 & 46.12 & 94.67 / 32.67 \\
\hline
\multirow{4}{*}{SVHN} &
\multicolumn{1}{c}{\textbf{Baseline}} & & &
70.76 / 0.13 & & & 
70.76 / 0 \\ 
\cmidrule{2-8}
 & FMNIST &  78.64 / 70.10 & 47.84 & 68.00 / 34.01 & 87.25 / 58.49 & 50.83 & 70.28 / 33.10  \\ 
\cmidrule{2-8}
 & CIFAR10 &  80.78 / 83.76 & 47.84 & 68.11 / 38.08 & 87.16 / 74.25 & 50.83 & 70.15 / 46.09 \\ 
\hline
\end{tabular}
}
\end{table*}

In this subsection, we evaluate the effectiveness of our data-free adversarial defense approach in scenarios where labeled data is limited, specifically when models are trained using semi-supervised learning (SSL) techniques. The goal is to determine the usefulness of our method in protecting models created under these settings. \\
We train target models using two State-of-the-art SSL methods: Flexmatch~\cite{Zhang2021FlexMatchBS} and Fixmatch~\cite{Sohn2020FixMatchSS}. We conduct our experiments on CIFAR10 and SVHN datasets, and throughout all the experiments, we utilize Wideresnet-16-1 (WRN) as the target model. For the target model trained on the CIFAR10 dataset, we use FMNIST and MNIST as arbitrary datasets, while for the models trained on the SVHN dataset, we use FMNIST and CIFAR10. We take the arbitrary model as Resnet18 for all these experiments. \\

\begin{table*}[hbp]
\centering
\caption{Comparison of the clean and adversarial accuracy of the target model trained using the SSL-Fixmatch method with and without DAD++ defense. The target Model architecture is WRN and Arbitrary Model architecture is Resnet18 across all experiments. The results show DAD++ is able to defend the target model trained using the SSL-Fixmatch method. Notations- TD.A:Target detector accuracy, Co.A:Correction accuracy, Cb.A:Combined accuracy.}
\label{SSL-Fixmatch}
\scalebox{0.8}{
\begin{tabular}{cccccccc}
\hline
\multirow[b]{3}{*}{\textbf{\begin{tabular}[c]{@{}c@{}}Target\\ Dataset\end{tabular}}} &
\multirow[b]{3}{*}{\textbf{\begin{tabular}[c]{@{}c@{}}Arbitrary\\ Dataset\end{tabular}}} &
\multicolumn{6}{c}{\textbf{Accuracy}} \\ \cmidrule{3-8}
& & 
\multicolumn{3}{c}{\textbf{PGD}} & 
\multicolumn{3}{c}{\textbf{Auto-Attack}} \\ \cmidrule{3-8} 
& &
\textbf{TD.A} &
\textbf{Co.A} &
\textbf{Cb.A} &
\textbf{TD.A} &
\textbf{Co.A} &
\textbf{Cb.A} \\ \hline \hline 
\multirow{4}{*}{CIFAR10} &
\multicolumn{1}{c}{\textbf{Baseline}} & & &
92.69 / 0 & & & 
92.69 / 0 \\ 
\cmidrule{2-8}
 & FMNIST &  91.89 / 71.02 & 49.05 & 91.45 / 29.58 & 87.33 / 82.44 & 49.29 & 89.17 / 37.55 \\
\cmidrule{2-8}
 & MNIST & 90.16 / 71.17 & 49.05 & 90.81 / 28.60 & 95.38 / 71.41 & 49.29 & 92.14 / 29.53 \\
\hline
\multirow{4}{*}{SVHN} &
\multicolumn{1}{c}{\textbf{Baseline}} & & &
94.52 / 0.96 & & & 
94.52 / 0.01 \\ 
\cmidrule{2-8}
& FMNIST &  84.65 / 61.13 & 76.97 & 94.46 / 28.54 & 86.93 / 73.95 & 78.89 & 94.28 / 55.52  \\ 
\cmidrule{2-8}
 & CIFAR10 & 74.10 / 77.54 & 76.97 & 94.24 / 52.52 & 90.44 / 87.28 & 78.89 & 94.28 / 71.19 \\ 
\hline
\end{tabular}}
\end{table*}

\textbf{Flexmatch: }The results for the Flexmatch method are reported in Table~\ref{SSL-Flexmatch}. For the CIFAR10 dataset, when evaluated against the PGD attack, we obtained impressive performance with our defense, irrespective of the arbitrary dataset. Whereas, against Auto attack, we observed the MNIST dataset to be a better choice for the arbitrary dataset. Overall, we observe clean accuracy drops by a maximum of $\approx$3\% across attacks (when compared to the baseline). while the adversarial accuracy improves by around 32-41\% across attacks. Similarly, for the SVHN dataset, we observe a minimal drop in clean accuracy of about 2\% and 1\% while defending against the PGD and Auto-attack, respectively. Here, the adversarial accuracy increases by $\approx$33-46\% across different attacks.

\textbf{Fixmatch: } We present the results for the experiments on the model trained using the Fixmatch method in Table~\ref{SSL-Fixmatch}. For CIFAR10 as the target dataset, we observe that using both FMNIST and MNIST as arbitrary dataset ensure good performance, but with FMNIST, slightly better performance is observed. Against the PGD attack, we observe a drop of $\approx$2\% in clean accuracy and an improvement of $\approx28-29\%$ in adversarial accuracy compared to the baseline. Against Auto attack, with a drop of around $\approx$3\% in clean accuracy, the adversarial accuracy improves by $\approx$37.5\% on using FMNIST as the arbitrary dataset. For the SVHN dataset, when CIFAR10 is an arbitrary dataset, we observe the clean accuracy drops by less than 1\%, and adversarial accuracy improves by more than 52\% and 70\% against PGD and Auto-attack, respectively.

The extensive experimentation with a variety of combinations of target and arbitrary models and datasets demonstrates that the DAD++ can effectively protect models trained using self-supervised learning techniques. In these experiments, the arbitrary model (Resnet18) is different from the target model (WideResNet); however, this does not have a negative impact on the performance of the DAD++ defense. Similar to subsection ~\ref{sl}, in this section also we observe that the target detector accuracy is dependent on the choice of arbitrary dataset.
\subsection{Data-free Knowledge Distillation}\label{kd}

\begin{table*}[t]
\centering
\caption{Comparison of the performance of the target model trained using the DAFL method with and without DAD++ defense. The results show that the use of DAD++ defense results in a significant increase in adversarial accuracy, while there is only a small decrease in clean accuracy.Resnet18 is both target model and arbitrary model architecture. Notations- TD.A:Target detector accuracy, Co.A:Correction accuracy, Cb.A:Combined accuracy.}
\label{KD-DAFL}
\scalebox{0.8}{
\begin{tabular}{cccccccc}
\hline
\multirow[b]{3}{*}{\textbf{\begin{tabular}[c]{@{}c@{}}Target\\ Dataset\end{tabular}}} &
\multirow[b]{3}{*}{\textbf{\begin{tabular}[c]{@{}c@{}}Arbitrary\\ Dataset\end{tabular}}} &
\multicolumn{6}{c}{\textbf{Accuracy}} \\ \cmidrule{3-8}
& & 
\multicolumn{3}{c}{\textbf{PGD}} & 
\multicolumn{3}{c}{\textbf{Auto-Attack}} \\ \cmidrule{3-8} 
& & 
\textbf{TD.A} &
\textbf{Co.A} &
\textbf{Cb.A} &
\textbf{TD.A} &
\textbf{Co.A} &
\textbf{Cb.A} \\ \hline \hline 
\multirow{4}{*}{CIFAR10} &

\multicolumn{1}{c}{\textbf{Baseline}} & & &
90.41 / 0 & & & 
90.41 / 0 \\ 
\cmidrule{2-8}
& FMNIST  & 94.42 / 91.36 & 39.79 & 89.48 / 33.64 & 97.5 / 83.28 & 38.8 & 90.35 / 30.27 \\
\cmidrule{2-8}
& MNIST  & 97.65 / 83.23 & 39.79 & 90.21 / 27.81 & 99.55 / 87.69 & 38.8 & 90.32 / 34.08 \\
\hline
\multirow{4}{*}{SVHN} &

\multicolumn{1}{c}{\textbf{Baseline}} & & &
94.41 / 3.33 & & & 
94.41 / 0.87 \\ 
\cmidrule{2-8}
& FMNIST  & 78.80 / 69.18 & 70.53 & 93.83 / 33.08 & 83.12 / 63.77 & 73.97 & 94.01 / 28.25  \\ 
\cmidrule{2-8}
& CIFAR10  & 75.51 / 76.53 & 70.53 & 92.67 / 49.22 & 89.12 / 73.73 & 73.97 & 93.92 / 52.49  \\ 
\hline
\end{tabular}
}
\end{table*}

\begin{table*}[b]
\centering
\caption{ \footnotesize Performance analysis of the target model trained using the ZSKT method with and without DAD++ defense. The target model architecture is WideResnet and the arbitrary model architecture is Resnet18. Integrating DAD++ into the ZSKT method shows a desirable performance boost over the baseline setting. Notations- TD.A:Target detector accuracy, Co.A:Correction accuracy, Cb.A:Combined accuracy.}
\label{KD-ZSKT}
\scalebox{0.8}{
\begin{tabular}{cccccccc}
\hline
\multirow{3}{*}{\textbf{\begin{tabular}[c]{@{}c@{}}Target\\ Dataset\end{tabular}}} &
\multirow{3}{*}{\textbf{\begin{tabular}[c]{@{}c@{}}Arbitrary\\ Dataset\end{tabular}}} &
\multicolumn{6}{c}{\textbf{Accuracy}} \\ \cmidrule{3-8}
& &
\multicolumn{3}{c}{\textbf{PGD}} & 
\multicolumn{3}{c}{\textbf{Auto-Attack}} \\ \cmidrule{3-8} 
& &
\textbf{TD.A} &
\textbf{Co.A} &
\textbf{Cb.A} &
\textbf{TD.A} &
\textbf{Co.A} &
\textbf{Cb.A} \\ \hline \hline 
\multirow{4}{*}{CIFAR10} &
\multicolumn{1}{c}{\textbf{Baseline}} & & &
84.26 / 0 & & & 
84.26 / 0 \\ 
\cmidrule{2-8}
 & FMNIST & 83.64/93.09 & 25.75 & 75.88/22.27 & 82.70/82.10 & 27.29 & 75.02/23.13 \\
\cmidrule{2-8}
& MNIST & 92.71/83.34 & 25.75 & 82.27/18.96 & 92.00 / 80.32 & 27.29 & 80.80 / 22.29 \\
\hline
\multirow{4}{*}{SVHN} &
\multicolumn{1}{c}{\textbf{Baseline}} & & &
94.37 / 0.69 & & & 
94.37 / 0.42 \\ 
\cmidrule{2-8}
& FMNIST & 74.36 / 69.15 & 67.97 & 93.72 / 33.36 & 66.82 / 76.65 & 71.37 & 90.55 / 51.21  \\ 
\cmidrule{2-8}
& CIFAR10 & 72.81 / 81.30 & 67.97 & 92.46 / 50.20 & 77.52 / 80.87 & 71.37 & 93.08 / 55.81  \\ 
\hline
\end{tabular}
}

\end{table*}

In this subsection, we study the efficacy of DAD++ in defending the models trained using data-free knowledge distillation (DFKD) methods. DFKD methods are useful when there is no training data available for training the model; instead, a pre-trained teacher model is available. However, the adversarial robustness of such models becomes an issue due to the unavailability of training data. We intend to defend the student models trained using DFKD methods by utilizing our DAD++ mechanism. For this, we conduct experiments using two State-of-the-Art DFKD methods: 1) Data-Free Learning of Student Networks (DAFL)~\cite{chen2019data} and 2) Zero Shot Knowledge Transfer (ZSKT)~\cite{Micaelli2019ZeroshotKT}. 

\textbf{DAFL: }DAFL trains a generator to learn training data distribution using the teacher network as a discriminator. This generator is then utilized to produce images for distilling knowledge from teacher to student model. We consider the student network trained using the DAFL method on CIFAR10 and SVHN datasets
as the target model. The architecture of both the student network and the arbitrary model is Resnet18, and the arbitrary dataset used to train the arbitrary model are FMNIST and MNIST. 

From the results reported in Table~\ref{KD-DAFL} for the CIFAR10 dataset, we observe a minimal drop of $\approx$1\% in clean accuracy, while the adversarial accuracy improves significantly by 27-34\% across attacks. Similarly, for the SVHN dataset, we notice a minimal drop in clean accuracy while achieving an adversarial accuracy of 52\% for the auto attack. The detector performance for the SVHN dataset is poor compared to CIFAR10, which may be due to the fact that the adversarial attack on the SVHN dataset is not strong, resulting in adversarial and clean images being very close to each other in feature space, making it difficult to estimate the decision boundary for the detector. Stronger attacks are easier to detect than weak adversarial attacks, however, denoising the images highly perturbed using a strong adversarial attack is more difficult for the correction module than images perturbed using a comparatively weaker attack.

\begin{table*}[b]
\centering
\caption{Notations- A:Amazon, D:DSLR, W:Webcam, TD.A:Target detector accuracy, Co.A:Correction accuracy, Cb.A:Combined accuracy. Comparison of the accuracy of the target model trained using the SHOT method with and without DAD++ defense on Office-31 Dataset. The target model architecture is Resnet50. arbitrary model architecture is Resnet18}
\label{Shot-Office}
\scalebox{0.8}{
\begin{tabular}{cccccccc}
\hline
\multirow{3}{*}{\textbf{\begin{tabular}[c]{@{}c@{}}Target\\ Dataset\end{tabular}}} &
\multirow{3}{*}{\textbf{\begin{tabular}[c]{@{}c@{}}Arbitrary\\ Dataset\end{tabular}}} &
\multicolumn{6}{c}{\textbf{Accuracy}} \\ \cmidrule{3-8}
& & 
\multicolumn{3}{c}{\textbf{PGD}} & 
\multicolumn{3}{c}{\textbf{Auto-Attack}} \\ \cmidrule{3-8} 
& &
\textbf{TD.A} &
\textbf{Co.A} &
\textbf{Cb.A} &
\textbf{TD.A} &
\textbf{Co.A} &
\textbf{Cb.A} \\ \hline \hline 
\multirow{3}{*}{A$\rightarrow$D} &
\multicolumn{1}{c}{\textbf{Baseline}} & & &
93.57 / 0 & & & 
93.57 / 0 \\ 
\cmidrule{2-8}
& FMNIST & 81.92 / 91.36 & 62.44 & 87.55 / 54.22 & 99.79 / 93.37 & 56.22 & 93.57 / 54.02 \\
\cmidrule{2-8}
& CIFAR10 & 98.59 / 97.59 & 62.44 & 93.57 / 60.44 & 99.9 / 93.57 & 56.22 & 93.57 / 56.22 \\
\hline
\multirow{3}{*}{A$\rightarrow$W} &

\multicolumn{1}{c}{\textbf{Baseline}} & & &
90.94 / 0 & & & 
90.94 / 0 \\ 
\cmidrule{2-8}
& FMNIST & 88.05 / 96.98 & 62.51 & 85.16 / 59.25 & 96.72 / 91.82 & 63.52 & 90.94 / 60.75 \\
\cmidrule{2-8}
& CIFAR10 & 90.94 / 97.86 & 62.51 & 89.31 / 60.00 & 99.9 / 90.94 & 63.52 & 90.94 / 62.14 \\
\hline
\multirow{3}{*}{D$\rightarrow$A} &

\multicolumn{1}{c}{\textbf{Baseline}} & & &
74.68 / 0 & & & 
74.68 / 0 \\ 
\cmidrule{2-8}
& FMNIST & 91.26 / 94.35 & 45.33 & 73.27 / 41.64 & 98.97 / 74.90 & 41.99 & 74.68 / 40.11 \\
\cmidrule{2-8}
& CIFAR10 & 96.80 / 92.08 & 45.33 & 74.33 / 40.75 & 99.9 / 74.68 & 41.99 & 74.68 / 38.37 \\
\hline
\multirow{3}{*}{D$\rightarrow$W} &

\multicolumn{1}{c}{\textbf{Baseline}} & & &
97.23 / 0 & & & 
97.23 / 0 \\ 
\cmidrule{2-8}
& FMNIST & 90.94 / 95.72 & 62.76 & 92.83 / 59.50 & 99.24 / 91.57 & 64.77 & 97.23 / 58.87 \\
\cmidrule{2-8}
& CIFAR10 & 96.72 / 97.10 & 62.76 & 96.60 / 61.38 & 99.9 / 97.23 & 64.77 & 97.23 / 62.89 \\
\hline
\multirow{3}{*}{W$\rightarrow$A} &

\multicolumn{1}{c}{\textbf{Baseline}} & & &
76.00 / 0 & & & 
76.00 / 0 \\ 
\cmidrule{2-8}
& FMNIST & 88.28 / 94.63 & 47.88 & 72.10 / 44.76 & 98.29 / 75.47 & 46.25 & 76.00 / 44.20 \\
\cmidrule{2-8}
& CIFAR10 & 95.63 / 94.53 & 47.88 & 75.40 / 44.73 & 99.9 / 76.00 & 46.25 & 76.00 / 44.52 \\
\hline
\multirow{3}{*}{W$\rightarrow$D} &

\multicolumn{1}{c}{\textbf{Baseline}} & & &
99.79 / 0 & & & 
99.79 / 0 \\ 
\cmidrule{2-8}
& FMNIST & 72.08 / 92.77 & 60.24 & 91.77 / 58.63 & 99.79 / 98.19 & 55.22 & 99.79 / 53.82 \\
\cmidrule{2-8}
& CIFAR10 & 98.99 / 97.59 & 60.24 & 99.80 / 60.10 & 99.9 / 99.79 & 55.22 & 99.79 / 54.82 \\
\hline
\end{tabular}
}

\end{table*}

\textbf{ZSKT: } Similar to DAFL, ZSKT utilizes a generator to create pseudo-training data for training the student model. For our DAD++ experiments, we use the pre-trained student models distilled using the ZSKT method. The target model's architecture is WideResnet-16-1. The results are reported in Table~\ref{KD-ZSKT}

For the CIFAR10 dataset, we notice that the clean accuracy decreases by $\approx$8-9\% compared to the baseline when we use FMNIST as an arbitrary dataset. In contrast, using the MNIST dataset as the arbitrary dataset resulted in a drop in clean accuracy of about 2-4\%. However, the adversarial accuracy was higher with FMNIST dataset compared to the MNIST dataset. This decrease in clean accuracy can be attributed to the lower detection accuracy on clean samples (around 82-83\%).
For the SVHN dataset, we use FMNIST and CIFAR10 as arbitrary datasets. In both cases, we observe a small drop (around 1-4\%) in clean accuracy compared to the baseline, across all the attacks. The detection accuracy is higher when the CIFAR10 dataset is an arbitrary dataset compared to the FMNIST dataset. Therefore, we get better adversarial accuracy with the CIFAR10 dataset than with the FMNIST dataset.

Overall, our method appears to be effective at providing defense for models created using a data-free knowledge distillation setup, where no training data is available.
\vspace{-0.09in}
\subsection{Source-free Domain Adaptation}\label{uda}

\begin{table*}[t]
\centering
\caption{ Notations- S:SVHN, M:MNIST, U:USPS, TD.A:Target detector accuracy, Co.A:Correction accuracy, Cb.A:Combined accuracy. Comparison of the accuracy of the target model trained using the SHOT method with and without DAD++ defense on Digit Dataset. The target model architecture is DTN for S$\rightarrow$M, and Resnet50 for M$\rightarrow$U. The arbitrary model architecture is Resnet18.}
\label{Shot-Digit}
\scalebox{0.8}{
\begin{tabular}{cccccccc}
\hline
\multirow{3}{*}{\textbf{\begin{tabular}[c]{@{}c@{}}Target\\ Dataset\end{tabular}}} &
\multirow{3}{*}{\textbf{\begin{tabular}[c]{@{}c@{}}Arbitrary\\ Dataset\end{tabular}}} &
\multicolumn{6}{c}{\textbf{Accuracy}} \\ \cmidrule{3-8}
& & 
\multicolumn{3}{c}{\textbf{PGD}} & 
\multicolumn{3}{c}{\textbf{Auto-Attack}} \\ \cmidrule{3-8} 
& & 
\textbf{TD.A} &
\textbf{Co.A} &
\textbf{Cb.A} &
\textbf{TD.A} &
\textbf{Co.A} &
\textbf{Cb.A} \\ \hline \hline 
\multirow{3}{*}{S$\rightarrow$M} &
\multicolumn{1}{c}{\textbf{Baseline}} & & &
99.02 / 0 & & & 
99.02 / 0 \\ 
\cmidrule{2-8}
& FMNIST & 91.32 / 79.01 & 43.45 & 99.01 / 23.97 & 97.61 / 74.76 & 44.26 & 99.02 / 22.82 \\
\cmidrule{2-8}
& CIFAR10 & 96.77 / 96.57 & 42.91 & 98.91 / 37.96 & 98.72 / 98.32 & 43.94 & 98.97 / 42.21 \\
\hline
\multirow{3}{*}{M$\rightarrow$U} &
\multicolumn{1}{c}{\textbf{Baseline}} & & &
98.17 / 0 & & & 
98.17 / 0 \\ 
\cmidrule{2-8}
& FMNIST & 74.19 / 86.50 & 38.01 & 97.96 / 25.43 & 82.25 / 81.50 & 28.38 & 98.01 / 22.15 \\
\cmidrule{2-8}
& CIFAR10 & 98.06 / 94.46 & 38.01 & 98.17 / 30.97 & 99.08 / 98.27 & 29.73 & 98.17 / 29.46 \\
\hline
\end{tabular}
}

\end{table*}

\begin{table*}[b]
\centering
\caption{Performance analysis of the target model trained using the DINE method with and without DAD++ defense on Office-31 dataset. The target Model architecture is Resnet50. arbitrary model architecture is Resnet18. DAD++ when integrated with the target model shows a significant increment in the adversarial performance with a negligible drop in clean accuracy. Notations- TD.A:Target detector accuracy, Co.A:Correction accuracy, Cb.A:Combined accuracy.}
\label{Dine-Office}
\scalebox{0.8}{
\begin{tabular}{cccccccc}
\hline
\multirow{3}{*}{\textbf{\begin{tabular}[c]{@{}c@{}}Target\\ Dataset\end{tabular}}} &
\multirow{3}{*}{\textbf{\begin{tabular}[c]{@{}c@{}}Arbitrary\\ Dataset\end{tabular}}} &
\multicolumn{6}{c}{\textbf{Accuracy}} \\ \cmidrule{3-8}
& &
\multicolumn{3}{c}{\textbf{PGD}} & 
\multicolumn{3}{c}{\textbf{Auto-Attack}} \\ \cmidrule{3-8} 
& &
\textbf{TD.A} &
\textbf{Co.A} &
\textbf{Cb.A} &
\textbf{TD.A} &
\textbf{Co.A} &
\textbf{Cb.A} \\ \hline \hline 
\multirow{3}{*}{A$\rightarrow$D} &

\multicolumn{1}{c}{\textbf{Baseline}} & & &
93.37 / 0 & & & 
93.37 / 0 \\ 
\cmidrule{2-8}
& FMNIST  & 98.79 / 99.79 & 54.41 & 92.97 / 54.41 & 91.96 / 91.36 & 59.83 & 93.17 / 56.22 \\
\cmidrule{2-8}
& CIFAR10  & 100 / 99.79 & 54.41 & 93.17 / 54.22 & 100 / 93.37 & 59.83 & 93.37 / 57.63\\
\hline
\multirow{3}{*}{A$\rightarrow$W} &

\multicolumn{1}{c}{\textbf{Baseline}} & & &
87.04 / 0 & & & 
87.04 / 0 \\ 
\cmidrule{2-8}
& FMNIST  & 91.69 / 92.32 & 56.22 & 85.03 / 50.31 & 96.79 / 87.16 & 58.11 & 86.16 / 57.48 \\
\cmidrule{2-8}
& CIFAR10  & 100 / 100 & 56.22 & 87.04 / 55.97 & 100 / 87.04 & 58.11 & 87.04 / 57.48 \\
\hline
\multirow{3}{*}{D$\rightarrow$A} &

\multicolumn{1}{c}{\textbf{Baseline}} & & &
70.85 / 0 & & & 
70.85 / 0 \\ 
\cmidrule{2-8}
& FMNIST  & 97.87 / 97.33 & 43.52 & 70.04 / 41.89 & 93.46 / 70.78 & 43.48 & 70.25 / 40.15 \\
\cmidrule{2-8}
& CIFAR10  & 99.96 / 99.14 & 43.52 & 70.85 / 42.53 & 99.92 / 70.89 & 43.48 & 70.82 / 40.89 \\
\hline
\multirow{3}{*}{D$\rightarrow$W} &

\multicolumn{1}{c}{\textbf{Baseline}} & & &
97.10 / 0 & & & 
97.10 / 0 \\ 
\cmidrule{2-8}
& FMNIST  & 96.10 / 93.58 & 58.86 & 96.35 / 53.58 & 99.87 / 94.96 & 65.28 & 97.10 / 60.75 \\
\cmidrule{2-8}
& CIFAR10  & 100 / 99.49 & 58.86 & 97.11 / 58.24 & 100 / 97.10 & 65.28 & 97.10 / 64.78 \\
\hline
\multirow{3}{*}{W$\rightarrow$A} &

\multicolumn{1}{c}{\textbf{Baseline}} & & &
71.13 / 0 & & & 
71.13 / 0 \\ 
\cmidrule{2-8}
& FMNIST  & 94.95 / 96.69 & 42.59 & 70.68 / 39.79 & 92.58 / 72.55 & 42.1 & 69.40 / 39.65 \\
\cmidrule{2-8}
& CIFAR10  & 99.92 / 99.57 & 42.59 & 71.13 / 42.71 & 99.92 / 71.24 & 42.1 & 71.13 / 39.30 \\
\hline
\multirow{3}{*}{W$\rightarrow$D} &

\multicolumn{1}{c}{\textbf{Baseline}} & & &
98.79 / 0 & & & 
98.79 / 0 \\ 
\cmidrule{2-8}
& FMNIST  & 90.56 / 97.18 & 50.8 & 97.19 / 48.39 & 100 / 98.39 & 59.03 & 98.79 / 58.8 \\
\cmidrule{2-8}
& CIFAR10  & 100 / 100 & 50.8 & 98.79 / 50.8 & 100 / 98.79 & 59.03 & 98.79 / 58.24 \\
\hline
\end{tabular}
}
\end{table*}

As discussed in section~\ref{sec:intro}, our proposed defense mechanism, DAD++, is capable of defending a target model created using source-free domain adaptation methods because it does not require access to the training data. This subsection examines the performance of DAD++ in defending the target models that were adapted from pre-trained source models using source-free domain adaptation techniques. We conduct experiments using three State-of-the-Art Source-free Domain Adaptation methods: SHOT~\cite{Liang2020DoWR}, DINE~\cite{Liang2021DINEDA}, and DECISION~\cite{Ahmed2021UnsupervisedMD}. SHOT is an unsupervised domain adaptation method that adapts a network to the target domain without access to the source data used to train the source model. Dine adapts the source model in a black box setup, where only the network predictions from the source model are available. Similar to SHOT, it also assumes the unavailability of the source data. DECISION is a domain adaptation method that adapts a network to the target domain using multiple source networks. \\

\textbf{SHOT: } Tables~\ref{Shot-Office} and~\ref{Shot-Digit} present the results with and without our DAD++ approach for the target model trained using the SHOT method on the Office-31 and Digit datasets. For the Office-31 dataset, there are six combinations of the source to target datasets: A$\rightarrow$D, A$\rightarrow$W, D$\rightarrow$A, D$\rightarrow$W, W$\rightarrow$A, and W$\rightarrow$D, where A stands for Amazon, D stands for DSLR, and W stands for Webcam. We consider the architecture of the target and arbitrary model to be Resnet50 and Resnet18, respectively.  We experiment with two different arbitrary datasets, CIFAR10 and FMNIST. 
When FMNIST is used as the arbitrary dataset, compared to the baseline, the clean accuracy drops by $\approx$4-8\% against the PGD attack, while against the auto attack, there is a minimal drop of around 0-1\% drop in clean accuracy. With CIFAR10 as the arbitrary dataset, performance with our DAD++ defense further improves against both the PGD and auto attacks with little or no drop in clean accuracy while the combined adversarial accuracy also does not decrease significantly compared to correction accuracy when CIFAR10 is the arbitrary dataset.

We evaluate the performance of our DAD++ method on the Digit dataset using two combinations of source-to-target datasets: S$\rightarrow$M and M$\rightarrow$U, where S, M, and U represent SVHN, MNIST, and USPS~\cite{hull1994database}, respectively. For the S$\rightarrow$M combination, the target model architecture is DTN~\cite{Zhang2015DeepTN}, and for the M$\rightarrow$U combination, it is Resnet50.

For S$\rightarrow$M, we observe across all attacks, there is a minimal drop in clean accuracy compared to the baseline. With CIFAR10 as the arbitrary dataset, the adversarial accuracy increases to 37.09\% against the PGD attack and 42.21\% against the Auto attack. For M$\rightarrow$U, we observe less than a 1\% drop in clean accuracy across attacks, while the adversarial accuracy increases by $\approx$29-30\% against both the PGD and Auto attacks when using the CIFAR10 dataset as the arbitrary dataset. \\

\begin{table*}[b]
\centering
\caption{Notations- A:Art,  C:Clipart, R:Real world, P: Product, TD.A:Target detector accuracy, Co.A:Correction accuracy, Cb.A:Combined accuracy.
Comparative analysis of target model's performance trained using the DINE method with and without DAD++ defense on Office Home dataset. The target Model architecture is Resnet50 and the arbitrary model architecture is Resnet18.}
\label{Dine-Office Home}
\scalebox{0.8}{
\begin{tabular}{cccccccc}
\hline
\multirow{3}{*}{\textbf{\begin{tabular}[c]{@{}c@{}}Target\\ Dataset\end{tabular}}} &
\multirow{3}{*}{\textbf{\begin{tabular}[c]{@{}c@{}}Arbitrary\\ Dataset\end{tabular}}} &
\multicolumn{6}{c}{\textbf{Accuracy}} \\ \cmidrule{3-8}
& & 
\multicolumn{3}{c}{\textbf{PGD}} & 
\multicolumn{3}{c}{\textbf{Auto-Attack}} \\ \cmidrule{3-8} 
& & 
\textbf{TD.A} &
\textbf{Co.A} &
\textbf{Cb.A} &
\textbf{TD.A} &
\textbf{Co.A} &
\textbf{Cb.A} \\ \hline \hline 
\multirow{3}{*}{A$\rightarrow$C} &

\multicolumn{1}{c}{\textbf{Baseline}}& & &
52.31 / 0 & & & 
52.31 / 0 \\ 
\cmidrule{2-8}
& FMNIST& 71.54 / 87.94 & 39.42 & 47.79 / 32.19 & 74.41 / 64.26 & 39.08 & 50.40 / 34.43 \\
\cmidrule{2-8}
& CIFAR10& 86.75 / 85.47 & 39.42 & 51.34 / 30.68 & 98.85 / 52.85 & 39.08 & 52.33 / 34.68\\
\hline
\multirow{3}{*}{A$\rightarrow$P} &

\multicolumn{1}{c}{\textbf{Baseline}}& & &
77.81 / 0 & & & 
77.81 / 0 \\ 
\cmidrule{2-8}
& FMNIST& 79.45 / 91.84 & 53.34 & 69.14 / 46.90 & 95.83 / 77.69 & 48.79 & 77.09 / 45.35 \\
\cmidrule{2-8}
& CIFAR10& 96.12 / 92.92 & 53.34 & 77.40 / 48.82 & 99.84 / 77.78 & 49.02 & 77.81 / 46.63 \\
\hline
\multirow{3}{*}{A$\rightarrow$R} &

\multicolumn{1}{c}{\textbf{Baseline}}& & &
81.45 / 0 & & & 
81.45 / 0 \\ 
\cmidrule{2-8}
& FMNIST& 79.84 / 91.98 & 61.71 & 74.59 / 54.44 & 95.70 / 79.66 & 58.06 & 81.02 / 53.06 \\
\cmidrule{2-8}
& CIFAR10& 95.89 / 94.83 & 61.71 & 81.13 / 57.24 & 99.97 / 80.99 & 58.29 & 81.46 / 54.83 \\
\hline
\multirow{3}{*}{C$\rightarrow$A} &

\multicolumn{1}{c}{\textbf{Baseline}}& & &
66.04 / 0 & & & 
66.04 / 0 \\ 
\cmidrule{2-8}
& FMNIST& 84.54 / 91.88 & 43.22 & 62.46 / 37.74 & 89.98 / 67.36 & 40.91 & 64.11 / 37.49 \\
\cmidrule{2-8}
& CIFAR10& 96.04 / 94.76 & 43.3 & 65.51 / 40.30 & 99.95 / 66.04 & 41.36 & 66.05 / 39.31 \\
\hline
\multirow{3}{*}{C$\rightarrow$P} &

\multicolumn{1}{c}{\textbf{Baseline}}& & &
76.88 / 0 & & & 
76.88 / 0 \\ 
\cmidrule{2-8}
& FMNIST& 80.80 / 90.94 & 51.16 & 69.50 / 45.10 & 89.68 / 78.89 & 46.72 & 75.26 / 44.36 \\
\cmidrule{2-8}
& CIFAR10& 91.91 / 90.85 & 51.16 & 75.90 / 45.19 & 99.97 / 76.90 & 47.15 & 76.89 / 45.01 \\
\hline
\multirow{3}{*}{C$\rightarrow$R} &

\multicolumn{1}{c}{\textbf{Baseline}}& & &
78.47 / 0 & & & 
78.47 / 0 \\ 
\cmidrule{2-8}
& FMNIST& 84.41 / 91.16 & 55.63 & 72.94 / 48.80 & 98.25 / 77.82 & 51.52 & 78.26 / 48.18 \\
\cmidrule{2-8}
& CIFAR10& 96.23 / 93.73 & 55.63 & 77.53 / 50.20 & 99.97 / 78.35 & 51.52 & 78.45 / 49.00 \\
\hline
\multirow{3}{*}{P$\rightarrow$A} &

\multicolumn{1}{c}{\textbf{Baseline}}& & &
78.47 / 0 & & & 
78.47 / 0 \\ 
\cmidrule{2-8}
& FMNIST& 84.71 / 92.62 & 42.43 & 60.65 / 37.74 & 83.93 / 68.39 & 39.51 & 63.16 / 34.73 \\
\cmidrule{2-8}
& CIFAR10& 96.00 / 94.76 & 42.43 & 63.70 / 37.87 & 99.71 / 64.52 & 40.13 & 64.32 / 37.66 \\
\hline
\multirow{3}{*}{P$\rightarrow$C} &

\multicolumn{1}{c}{\textbf{Baseline}}& & &
64.31 / 0 & & & 
64.31 / 0 \\ 
\cmidrule{2-8}
& FMNIST& 78.16 / 89.62 & 36.42 & 45.68 / 30.70 & 78.32 / 60.54 & 34.57 & 47.33 / 30.38 \\
\cmidrule{2-8}
& CIFAR10& 90.53 / 88.04 & 36.56 & 48.41 / 29.78 & 97.98 / 50.08 & 34.13 & 48.98 / 30.08 \\
\hline
\multirow{3}{*}{P$\rightarrow$R} &

\multicolumn{1}{c}{\textbf{Baseline}}& & &
82.25 / 0.16 & & & 
82.25 / 0 \\ 
\cmidrule{2-8}
& FMNIST& 80.28 / 90.19 & 60.11 & 74.89 / 52.56 & 81.23/89.32 &  57.34 & 75.87/52.56\\
\cmidrule{2-8}
& CIFAR10& 94.69 / 94.07 & 60.11 & 81.71 / 54.53 & 99.8 / 82.12 & 56.5 & 82.28 / 53.52 \\
\hline
\multirow{3}{*}{R$\rightarrow$A} &

\multicolumn{1}{c}{\textbf{Baseline}}& & &
69.83 / 0 & & & 
69.83 / 0 \\ 
\cmidrule{2-8}
& FMNIST& 87.06 / 94.56 & 45.07 & 65.88 / 41.24 & 96.95 / 70.53 & 40.58 & 69.59 / 38.85  \\
\cmidrule{2-8}
& CIFAR10& 95.05 / 94.39 & 45.07 & 69.06 / 42.07 & 99.91 / 69.79 & 41.86 & 69.84 / 39.51 \\
\hline
\multirow{3}{*}{R$\rightarrow$P} &

\multicolumn{1}{c}{\textbf{Baseline}}& & &
84.43 / 0 & & & 
84.43 / 0 \\ 
\cmidrule{2-8}
& FMNIST& 78.37 / 90.94 & 57.76 & 76.10 / 51.66 & 92.67 / 81.79 & 54.2 & 83.08 / 48.68 \\
\cmidrule{2-8}
& CIFAR10& 92.72 / 91.66 & 57.76 & 83.40 / 51.39 & 99.97 / 83.53 & 54.35 & 84.43 / 51.93 \\
\hline
\multirow{3}{*}{R$\rightarrow$C} &

\multicolumn{1}{c}{\textbf{Baseline}}& & &
57.06 / 0.16 & & & 
57.06 / 0 \\ 
\cmidrule{2-8}
& FMNIST& 77.20 / 86.30 & 42.03 & 51.96 / 33.29 & 82.10 / 65.36 & 41.05 & 55.23 / 36.79 \\
\cmidrule{2-8}
& CIFAR10& 87.51 / 86.73 & 42.03 & 55.67 / 33.33 & 98.51 / 57.47 & 40.52 & 57.04 / 36.33 \\
\hline
\end{tabular}
}

\end{table*}
\textbf{DINE: }Tables~\ref{Dine-Office} and~\ref{Dine-Office Home} present the results for the target model trained using the DINE method on the Office-31 and Office Home datasets, respectively. Office-31 dataset contains a subset of images from three domains: Amazon (A), DSLR (D), and Webcam (W), whereas the Office Home dataset contains images from four domains, which are Art (A), Clipart (C), Realworld (R) and Product(P). The target model architecture is Resnet50. Resnet18 is used as the arbitrary model, which is trained on two different arbitrary datasets, CIFAR10 and FMNIST. When FMNIST is used as the arbitrary dataset, we again observe a drop of $\approx$4-8\% clean accuracy against the PGD attack, while against the auto attack, we observed almost no drop in clean accuracy compared to the baseline. When CIFAR10 was used as the arbitrary dataset, our method performed even better against both PGD and auto attacks with little to no drop in clean accuracy. We observe similar results for the Office-31 dataset.

\textbf{DECISION: }Tables~\ref{Decision-Office} and~\ref{Decision-Office Home} present the results for the target model trained using the DECISION method on the Office-31 and Office Home datasets, respectively. As DECISION is a multi-source domain adaptation method, there are a total of four combinations of source to target for the Office Home dataset. The network is adapted to the target domain using the other three domains as source domains. Similarly, for the Office-31 dataset, there are a total of three combinations. We use Resnet50 as the target model architecture. Resnet18 is the arbitrary model trained using the arbitrary datasets CIFAR10 and FMNIST.

We observe, similar to previous experiments that CIFAR10 as an arbitrary dataset is better than FMNIST. We consistently observe that with CIFAR10, the clean accuracy drops by less than 1\%, while adversarial accuracy increases significantly compared to the baseline. With FMNIST, we observe a maximum of $\approx$4-5\% drop in clean accuracy across attacks. We get a significant improvement in adversarial accuracy with both CIFAR10 and FMNIST datasets, but compared to CIFAR10, the FMNIST dataset yields lesser combined clean and adversarial accuracy.

From the reported results, we make the following observations:
\begin{itemize}
    \item Our method provides a robust defense to target models, even when target dataset images have higher resolution than those in the arbitrary datasets (Arbitrary datasets CIFAR10 and FMNIST have a resolution of 32 and 28, respectively, while the target dataset Office-31 has a resolution of 256). 
    \item Our method exhibits strong performance even with a limited number of images in the target domain test set. For instance, the DSLR domain in the Office-31 dataset comprises only 498 images, yet our approach still effectively improves adversarial accuracy without decreasing clean accuracy substantially.
    \item We trained an arbitrary detector using only the PGD attack throughout all the experiments. Despite this, the adapted target detector demonstrates good performance against various adversarial attacks, such as AutoAttack and BIM.
\end{itemize}
\begin{table*}[htp]
\centering
\caption{Comparison of the accuracy of the target model trained using the DECISION method with and without DAD++ defense on Office-31 Dataset. Notations- TD.A:Target detector accuracy, Co.A:Correction accuracy, Cb.A:Combined accuracy.} 
\label{Decision-Office}
\scalebox{0.8}{
\begin{tabular}{cccccccc}
\hline
\multirow{3}{*}{\textbf{\begin{tabular}[c]{@{}c@{}}Target\\ Dataset\end{tabular}}} &
\multirow{3}{*}{\textbf{\begin{tabular}[c]{@{}c@{}}Arbitrary\\ Dataset\end{tabular}}} &
\multicolumn{6}{c}{\textbf{Accuracy}} \\ \cmidrule{3-8}
& & 
\multicolumn{3}{c}{\textbf{PGD}} & 
\multicolumn{3}{c}{\textbf{Auto-Attack}} \\ \cmidrule{3-8} 
& &
\textbf{TD.A} &
\textbf{Co.A} &
\textbf{Cb.A} &
\textbf{TD.A} &
\textbf{Co.A} &
\textbf{Cb.A} \\ \hline \hline 
\multirow{3}{*}{(D,W)$\rightarrow$A} &

\multicolumn{1}{c}{\textbf{Baseline}} & & &
74.86 / 0 & & & 
74.86 / 0 \\ 
\cmidrule{2-8}
 & FMNIST &  83.20 / 92.68 & 46.39 & 73.52 / 39.83 & 90.55 / 73.69 & 46.78 & 74.12 / 38.62 \\
\cmidrule{2-8}
& CIFAR10 &  98.08 / 98.68 & 46.39 & 74.46 / 45.58 & 97.55 / 75.04 & 46.5 & 74.69 / 44.66 \\
\hline
\multirow{3}{*}{(A,W)$\rightarrow$D} &

\multicolumn{1}{c}{\textbf{Baseline}} & & &
98.99 / 0 & & & 
98.99 / 0 \\ 
\cmidrule{2-8}
& FMNIST &  77.30 / 98.19 & 70.68 & 94.38 / 69.28 & 92.77 / 93.57 & 72.89 & 98.99 / 64.26 \\
\cmidrule{2-8}
 & CIFAR10 &  99.39 / 99.79 & 70.68 & 99.00 / 70.08 & 100 / 98.99 & 72.89 & 98.99 / 72.29 \\
\hline
\multirow{3}{*}{(A,D)$\rightarrow$W} &

\multicolumn{1}{c}{\textbf{Baseline}} & & &
98.11 / 0 & & & 
98.11 / 0 \\ 
\cmidrule{2-8}
 & FMNIST &  93.58 / 93.58 & 67.54 & 97.11 / 62.01 & 90.69 / 97.10 & 66.16 & 96.23 / 64.03 \\
\cmidrule{2-8}
 & CIFAR10 &  100 / 97.48 & 67.54 & 98.11 / 67.04 & 100 / 97.86 & 66.16 & 98.11 / 65.79 \\
\hline
\end{tabular}
}
\end{table*}

\begin{table*}[htp]
\centering
\caption{Comparison of the accuracy of the target model trained using the DECISION method with and without DAD++ defense on Office Home Dataset. Across all experiments, the target model is Resnet50 and the arbitrary model is Resnet18. Notations- TD.A:Target detector accuracy, Co.A:Correction accuracy, Cb.A:Combined accuracy.}
\label{Decision-Office Home}
\scalebox{0.8}{
\begin{tabular}{cccccccc}
\hline
\multirow{3}{*}{\textbf{\begin{tabular}[c]{@{}c@{}}Target\\ Dataset\end{tabular}}} &
\multirow{3}{*}{\textbf{\begin{tabular}[c]{@{}c@{}}Arbitrary\\ Dataset\end{tabular}}} &
\multicolumn{6}{c}{\textbf{Accuracy}} \\ \cmidrule{3-8}
& &
\multicolumn{3}{c}{\textbf{PGD}} &
\multicolumn{3}{c}{\textbf{Auto-Attack}} \\ \cmidrule{3-8} 
& & 
\textbf{TD.A} &
\textbf{Co.A} &
\textbf{Cb.A} &
\textbf{TD.A} &
\textbf{Co.A} &
\textbf{Cb.A} \\ \hline
\multirow{3}{*}{(C,P,R)$\rightarrow$A} &
\textbf{Baseline} & & &
74.49 / 0 & & & 
74.49 / 0 \\ 
\cmidrule{2-8}
& FMNIST & 85.16 / 95.42 & 51.00 & 71.08 / 46.52 & 81.78 / 78.65 & 53.68 & 69.22 / 50.43 \\
\cmidrule{2-8}
 & CIFAR10 & 96.78 / 97.69 & 51.00 & 74.37 / 48.58 & 99.54 / 74.61 & 53.68 & 74.50 / 50.89 \\
\hline
\multirow{3}{*}{(A,P,R)$\rightarrow$C} &

\textbf{Baseline} & & &
62.03 / 0.06 & & & 
62.03 / 0 \\ 
\cmidrule{2-8}
 & FMNIST &  77.38 / 88.77 & 46.52 & 56.45 / 37.18 & 81.21 / 60.20 & 46.66 & 61.35 / 32.71 \\
\cmidrule{2-8}
 & CIFAR10 &  89.07 / 89.75 & 46.52 & 61.08 / 37.09 & 87.03 / 64.44 & 46.66 & 60.76 / 41.70 \\
\hline
\multirow{3}{*}{(A,C,R)$\rightarrow$P} &

\textbf{Baseline} & & &
83.55 / 0 & & & 
83.55 / 0 \\ 
\cmidrule{2-8}
 & FMNIST &  94.75 / 71.18 & 51.45 & 82.56 / 32.69 & 88.12 / 84.61 & 52.71 & 80.09 / 50.33 \\
\cmidrule{2-8}
 & CIFAR10 &  95.06 / 97.29 & 51.45 & 82.86 / 51.12 & 98.19 / 83.73 & 52.71 & 83.51 / 51.88  \\
\hline
\multirow{3}{*}{(A,C,P)$\rightarrow$R} &

\textbf{Baseline} & & &
82.99 / 0 & & & 
82.99 / 0 \\ 
\cmidrule{2-8}
 & FMNIST &  87.37 / 96.21 & 57.14 & 78.66 / 53.87 & 96.69 / 80.44 & 59.55 & 82.56 / 55.27 \\
\cmidrule{2-8}
 & CIFAR10 &  98.46 / 97.01 & 57.14 & 82.72 / 53.96 & 99.49 / 82.00 & 59.55 & 82.99 / 56.55 \\
\hline
\end{tabular}
}
\end{table*}

\section{Conclusion}
In this paper, we presented a comprehensive test time detection and correction approach to achieve adversarial robustness in the absence of training data. We demonstrated the performance of each of our proposed modules, including detection and correction, through various experiments. The results across various adversarial attacks, datasets, applications and architectures showed the effectiveness of our method. The combined performance of the modules did not significantly compromise the clean accuracy, while achieving significant improvements in adversarial accuracy, even against state-of-the-art Auto Attack. Our data-free approach yielded competitive results when compared to data-dependent approaches. Additionally, we highlighted that our detection module is independent of the correction module, meaning any data-dependent detection approach can benefit from our correction module at test time to correct adversarial samples after successfully detecting them.

\section*{Funding/ conflict of interest}
\noindent This work is partially supported by a Young Scientist Research Award (Sanction no. 59/20/11/2020-BRNS) to Anirban Chakraborty from DAE-BRNS, India. 

Further, the authors have no financial or proprietary interests in any material discussed in this article.

\section*{Data availability statement}
\noindent The data that support the findings of this study are publicly available at \href{https://www.cs.toronto.edu/~kriz/cifar.html}{CIFAR10~\cite{krizhevsky2009learning}}, \href{https://www.kaggle.com/c/tiny-imagenet}{Tiny Imagenet~\cite{le2015tiny}} \href{https://www.vision.caltech.edu/datasets/cub_200_2011/}{CUB~\cite{WahCUB_200_2011}}, \href{https://github.com/zalandoresearch/fashion-mnist}{FMNIST~\cite{xiao2017fashion}}, \href{https://www.kaggle.com/datasets/hojjatk/mnist-dataset}{MNIST~\cite{lecun1998gradient}}, \href{http://ufldl.stanford.edu/housenumbers/}{SVHN~\cite{SVHN}}, \href{https://faculty.cc.gatech.edu/~judy/domainadapt/#datasets_code}{Office-31~\cite{saenko2010adapting}}, \href{https://www.kaggle.com/datasets/bistaumanga/usps-dataset}{USPS~\cite{hull1994database}}, and \href{https://www.hemanthdv.org/officeHomeDataset.html}{Office Home~\cite{venkateswara2017deep}}.

{\tiny
\bibliography{sn-bibliography}
}

\end{document}